\documentclass{article}

\usepackage[preprint]{neurips_2026}
\usepackage[utf8]{inputenc}
\usepackage[T1]{fontenc}
\usepackage{amsmath}
\usepackage{amsfonts}
\usepackage{array}
\usepackage{booktabs}
\usepackage{float}
\usepackage{graphicx}
\usepackage{hyperref}
\hypersetup{
  hidelinks,
  pdftitle={The Right Memory in the Wrong Context: Verifying Retrieval Admissibility in Long-Term Agent Memory},
  pdfauthor={Zi Wang, Xingqiao Wang, Emmanuel Addai, Devika Ambekar, Xiaowei Xu}
}
\usepackage{microtype}
\usepackage{nicefrac}
\usepackage{url}
\usepackage{xcolor}

\title{The Right Memory in the Wrong Context:\\
Verifying Retrieval Admissibility in Long-Term Agent Memory}

\author{%
  Zi Wang \quad Xingqiao Wang \quad Emmanuel Addai \\
  \textbf{Devika Ambekar \quad Xiaowei Xu}\thanks{Corresponding author.} \\
  University of Arkansas at Little Rock \\
  \texttt{zwang@ualr.edu} \quad
  \texttt{xwang1@ualr.edu} \quad
  \texttt{eaddai1@ualr.edu} \\
  \texttt{dmambekar@ualr.edu} \quad
  \texttt{xwxu@ualr.edu}
}

\begin{document}

\maketitle

\begin{abstract}
Agents equipped with long-term memory can retrieve relevant information that is inadmissible for the current request because it belongs to another principal, violates policy, or reflects an incompatible lifecycle state. Recall and final-answer accuracy do not reveal this: a route can appear safe by missing required evidence, while a correct answer may follow inadmissible prompt exposure. We introduce a \emph{retrieval-admissibility verification framework} that assigns each memory--query pair one of three statuses (admissible, inadmissible, or unresolved), compares routes at matched required-evidence recall with bounds for unresolved cases, and tracks memory IDs through prompt exposure while linking exposure to target-level disclosure. We evaluate its stages on separate, non-pooled populations. We conduct a
post-hoc top-20 reanalysis of frozen rankings from two public
long-term-memory benchmarks, RHELM and MemOps, covering \textbf{3,767}
queries. All released anchors lie within trusted query namespaces; with
within-namespace scores unchanged, off-namespace filtering cannot lower
their ranks. Top-20 anchor recall increases from \textbf{0.432 to 0.533},
$80\%$ recall feasibility from \textbf{0.237 to 0.311}, and exact
similarity evaluations decrease by \textbf{98.3\%}. In a frozen 72-case development diagnostic, a released-metadata reference preserves required evidence, whereas neither text-only verifier detects violations under the $1\%$ required-anchor false-denial limit. Across \textbf{1,523} paired benchmark-native cases, namespace routing is associated with judged-accuracy gains of \textbf{0.053--0.068} across three readers; recall also changes, so this comparison is observational. In 16 controlled exposure scenarios, only \textbf{one of four} reader-specific $95\%$ confidence intervals excludes zero for relevant-inadmissible literal disclosure (\(\mathbf{+0.156}\), $95\%$ CI $[0.031, 0.312]$). These results motivate separate verification of candidate support, admissibility, prompt exposure, and answer disclosure.
\end{abstract}

\section{Introduction}
\label{sec:introduction}

A consequential failure mode in long-term agent memory is retrieving an answer-bearing
record that is ineligible in the current context.  In audited public histories,
another principal's market note, an invalidated departure date, and a forgotten
battery detail all match their query topics; old medication states are instead
required when the query asks about change.  Recall and answer accuracy do not establish
whether this evidence may cross the prompt boundary for the current principal,
purpose, intent, and time.

We study this gap through \emph{retrieval admissibility}: whether a memory that is
relevant to a query is also eligible under the current principal, purpose, intent,
and time.  Relevance asks whether a record can answer the question; admissibility asks
whether it may be used to answer it.  The decision belongs to the memory--query
context, not the record alone: a superseded value may be inadmissible for
``what is true now?'' yet indispensable for ``how did it change?''  Likewise, a
topically matched record may remain unusable because it belongs to another principal,
violates an active policy, or was explicitly forgotten.

\begin{figure}[!t]
  \centering
  \includegraphics[width=\textwidth]{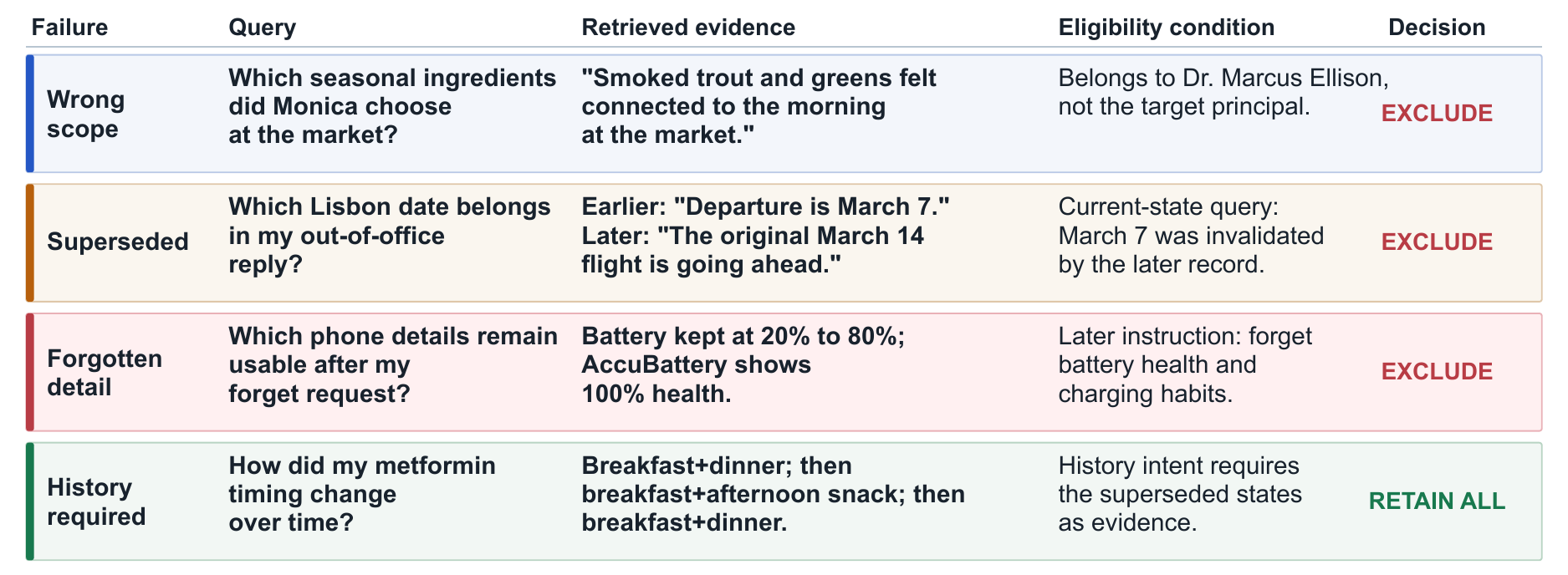}
  \caption{\textbf{Audited retrieval-admissibility cases.}
  Wrong-scope, superseded, and explicitly forgotten records are excluded, whereas
  superseded medication states are retained for a history query.  The first row is
  from RHELM and the remainder from MemOps; text is abridged, but verdicts are not.}
  \label{fig:examples}
\end{figure}

Figure~\ref{fig:examples} grounds these cases in their query, memory text, governing
condition, and decision.  We operationalize the resulting query-conditioned target
with matched-recall route accounting, record-level prompt exposure, and answer-level
disclosure.  Missing
governance evidence remains unresolved, and a route cannot appear safer merely by
returning less useful context.  The evaluation asks three stagewise verification
questions.  \textbf{RQ1 (candidate support):} how large is the trusted-provenance
reachability gain, and what do the controls explain?  \textbf{RQ2 (eligibility
decision):} can released fields or text-only verifiers reject known violations
without denying required evidence?  \textbf{RQ3 (boundary enforcement):} do route
changes reach answer behavior, and when does inadmissible exposure become disclosure?
Separate benchmark-native and controlled populations keep support, decision, and
reader failures identifiable.

We make three contributions: (i) a query-conditioned, three-valued admissibility
target with matched-recall partial-identification bounds; (ii) a protocol that tracks
memory IDs through prompt exposure and links exposure to target-level answer
disclosure, while separating scope, policy, and lifecycle violations; and (iii)
candidate-support, eligibility, and exposure studies. Trusted namespace restriction raises candidate
support and is associated with higher judged utility on the same frozen cases; text-only
gates over-deny required evidence; and one of four readers shows increased
literal-marker disclosure after relevant-inadmissible exposure. Code, released evidence derivatives, and reproduction scripts are available in our \href{https://github.com/ziwang11112/right-memory-wrong-context}{\textcolor{blue}{\underline{GitHub repository}}}.

\section{Related Work}

\paragraph{Long-term memory capability.}
LoCoMo, LongMemEval, MemBench, and MemoryAgentBench evaluate retention, retrieval,
reasoning, and capacity over long or multi-session histories
\citep{maharana2024locomo,wu2025longmemeval,tan2025membench,hu2026memoryagentbench}.  RHELM adds
heterogeneous temporal histories, and LongMemEval-V2 tests compact evidence gathering
from accumulated experience \citep{zhang2026rhelm,wu2026longmemevalv2}.

\paragraph{Validity, governance, and contextual use.}
MemOps, Memora, STALE, A-TMA, and LoCoMo-Plus expose conflicts among recall,
mutation, forgetting, outdated state, and latent constraints
\citep{hao2026memops,uddin2026memora,chao2026stale,shi2026atma,li2026locomoplus}.
Governed-memory work studies access control, contextual information flow,
provenance, forgetting, and supersession
\citep{ren2026gatemem,mireshghallah2026cimemories,fu2026ciwork,rezazadeh2025collaborative,
 margalit2026governed}.  GateMem benchmarks multi-principal governance, MemGate
predicts query-conditioned admission, and MRMS/MAP-Graph enforce eligibility or
provenance before generation
\citep{ren2026gatemem,zhang2026memgate,li2026mrms,wang2026mapgraph}.  Selection
Integrity diagnoses information-flow-blind accumulation; authorization and selective-
disclosure systems motivate pre-generation enforcement
\citep{fei2026selectionintegrity,hu2014abac,arceo2026securing,almasoud2026sdrag}.
We instead audit each retrieved record at matched recall, with unresolved-label bounds
and exposure-to-disclosure traces.

\paragraph{Stagewise analytical evaluation.}
HELM, AgentBoard, RAGChecker, and MEMPROBE motivate multi-metric, intermediate-stage
analysis \citep{liang2023helm,ma2024agentboard,ru2024ragchecker,ma2026memprobe}.
Building on that view, we audit scope, policy, and lifecycle eligibility at matched
evidence recall, retain the same memory ID through prompt exposure, and score
answer-level target disclosure rather than collapsing the path into answer accuracy.

\section{Retrieval-Admissibility Verification Framework}
\label{sec:framework}

\subsection{Framework overview}

\begin{figure}[!t]
  \centering
  \includegraphics[width=\textwidth]{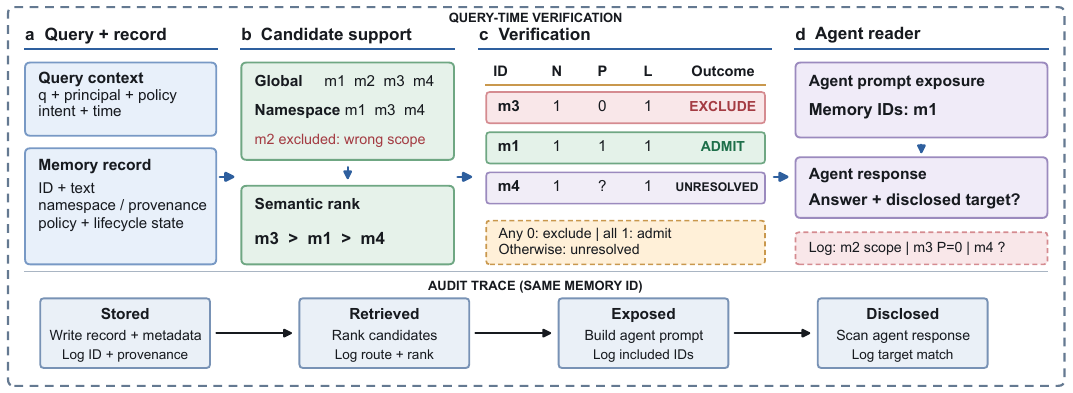}
  \caption{\textbf{Agent-memory retrieval-admissibility verification framework.}
  Query metadata constrain candidate support, semantic retrieval ranks IDs, and
  policy and lifecycle verification precede agent-prompt assembly.  The lower trace
  follows one memory ID from storage through retrieval and exposure, followed by
  target-level disclosure scoring; disclosure is observable output, not internal
  causal use.  Each study reports its population and reader-specific estimates
  separately.}
  \label{fig:framework}
\end{figure}

Let $\mathcal M$ be the stored memories.  Given a query context $z$, a candidate
memory $m\in\mathcal M$, and a retrieval arm $\alpha$,
the framework verifies three linked objects: record-level eligibility,
matched-recall route quality, and disclosure after candidate exposure to a
downstream agent reader.  It does not
prescribe one retrieval or filtering algorithm.  Instead, symbolic metadata, learned
verifiers, and unfiltered routes are evaluated against the same target and audit
path.  The common unit is one memory ID through retrieval and agent-prompt exposure;
the answer trace records only whether target content is disclosed in the agent
response.
Here a \emph{namespace} is a trusted provenance identifier---for example, an
authenticated account, persona, profile, tenant, or project---that defines candidate
support.  It is not a topic cluster or an identity inferred from semantic text.
Under complete trusted namespace labels and identical within-namespace scores, every
source-required anchor lies in namespace support. Namespace pre-filtering therefore
cannot worsen a released anchor's rank under these trusted-namespace assumptions.
We use it as a structural positive control; the empirical questions are effect
magnitude, exact search work, robustness to provenance errors, and route-to-reader
association.
Figure~\ref{fig:framework} summarizes the resulting support--verification--reader
pipeline and the ID-level audit trace beneath it.

\subsection{Query-conditioned admissibility}
\label{sec:query-conditioned-admissibility}

Let $z=(q,p,\kappa,i_q,t)$ contain query text $q$, authenticated principal $p$,
governing policy $\kappa$ (including purpose constraints), query intent $i_q$, and
evaluation time $t$.
For memory $m$, the evaluator
records relevance $r_z(m)$,
scope $n_z(m)$, policy permission $g_z(m)$, and lifecycle compatibility $\ell_z(m)$.
Each axis lies in $\mathcal{T}=\{1,0,?\}$: established positive, established
negative, or unresolved.  Here \emph{unresolved} names an evaluator state;
\emph{unknown} below is reserved for an explicit verifier abstention.

For $d$ labels, we combine three-valued decisions using
\begin{equation}
\bigwedge\nolimits_{3}(x_1,\ldots,x_d)=
\begin{cases}
0, & \exists j:x_j=0,\\
1, & \forall j:x_j=1,\\
?, & \text{otherwise}.
\end{cases}
\label{eq:three-valued}
\end{equation}
This is the strong Kleene conjunction on $\mathcal T$ \citep{kleene1952introduction}.
The diagnostic predicates need not be mutually exclusive: their conjunction defines
admissibility, while typed violation rates describe overlapping exclusion reasons
and must not be summed.
Using Equation~\ref{eq:three-valued}, the admissibility and usability targets for a
memory--context pair are
\begin{align}
a_z(m)
&=\bigwedge\nolimits_{3}
  \left(n_z(m),g_z(m),\ell_z(m)\right),
\label{eq:admissible}\\
u_z(m)
&=\bigwedge\nolimits_{3}
  \left(r_z(m),a_z(m)\right).
\label{eq:usable}
\end{align}

Equations~\ref{eq:admissible} and~\ref{eq:usable} separate governance eligibility
from relevance.  One established governance violation makes a candidate
inadmissible; usability also requires relevance.  Unresolved labels are never imputed.  Lifecycle remains
relational.  Let $s_m\in\{\mathrm{current},\mathrm{stale},
\mathrm{superseded},?\}$ be the released memory state and
$i_q\in\{\mathrm{current},\mathrm{history},?\}$ the query intent.  We set
$\ell_z(m)=1$ for current records; stale or superseded records receive 1 for history
intent and 0 for current intent; either unresolved input yields~$?$.  Thus an earlier
state can be required for a history query.  This released-field evaluator rule neither
asserts that every retained record is relevant nor infers latent intent.

\subsection{Stagewise record accounting}

The evaluation labels
$\mathbf{x}_z(m)=(n_z(m),g_z(m),\ell_z(m))$ are distinct from method-visible
metadata $\widetilde{\mathbf{x}}_{z\alpha}(m)$ and from a verifier prediction
$\widehat a_{zm\alpha}\in\{0,1,\mathrm{unknown}\}$.  Corruption changes only the
visible metadata; the evaluation
labels and evidence anchors remain fixed.  For binary indicators, $W_m=1$ means that
$m$ is stored, $R_{qm\alpha}=1$ that arm $\alpha$ returns it for $q$, and
$E_{qm\alpha}=1$ that prompt assembly exposes it to the downstream agent reader.
They obey
\begin{equation}
E_{qm\alpha}\leq R_{qm\alpha}\leq W_m.
\label{eq:stage-order}
\end{equation}
Equation~\ref{eq:stage-order} makes prompt exposure a subset of retrieval rather than
an alias for it.  We also record $Y_{q\alpha}=1$ when the prespecified scorer detects the target content
in the agent response.  Disclosure is observable output, not evidence of the agent's
internal causal use.  Thus the trace localizes failures in
candidate support, verification, prompt assembly, or disclosure.  For example,
trusted namespace support ranks only
$\mathcal{C}^{N}_{z\alpha}=\{m\in\mathcal{M}:\widetilde n_{z\alpha}(m)=1\}$; other arms substitute global,
governance-filtered, routed, or verifier-filtered support without changing the audit
unit.  Operational handling of missing information is arm-specific and declared:
strict namespace matching excludes a missing namespace, released policy and lifecycle
gates remove only established violations, and the abstaining text verifier declines
to delete candidates with high predicted unknown mass.

\subsection{Matched-recall route accounting}
\label{sec:matched-risk}

Let $\mathcal{G}_q$ be the source-released required evidence anchors used to score
recall.  These anchors are fixed independently of our admissibility evaluator: a
coarse released governance field may still flag an anchor in typed-risk accounting,
but it cannot silently redefine source gold.  A query with no released anchor is
recall-unevaluable.
Let $\pi_{q\alpha}=(m_1,\ldots,m_n)$ be the ranking from arm $\alpha$.  For a declared
budget $K$, $\pi_{q\alpha}$ denotes the route truncated to $n\leq K$; failure to reach
$\tau$ within that route is infeasibility.  At target recall $\tau$, we score the
smallest qualifying prefix:
\begin{align}
\operatorname{Rec}_{q\alpha}(k)
&=\frac{|\{m_1,\ldots,m_k\}\cap\mathcal{G}_q|}{|\mathcal{G}_q|},\\
k^*_{q\alpha}(\tau)
&=\min\{k:\operatorname{Rec}_{q\alpha}(k)\geq\tau\},\\
P_{q\alpha}(\tau)
&=\{m_1,\ldots,m_{k^*_{q\alpha}(\tau)}\}.
\label{eq:matched-prefix}
\end{align}
where $\min\emptyset=\infty$.  The route is feasible when
$F_{q\alpha}(\tau)=\mathbf{1}\{k^*_{q\alpha}(\tau)<\infty\}=1$.  Charging every item
before the same recall target prevents a route from appearing safer merely by
returning less useful evidence; Equation~\ref{eq:matched-prefix} is therefore the
common comparison point for every route.

For a feasible prefix of size $N_{q\alpha}=|P_{q\alpha}(\tau)|$, let
$c^A_{q\alpha}$, $v^A_{q\alpha}$, and $h^A_{q\alpha}$ count established admissibility violations,
established positives, and unresolved labels, so
$N_{q\alpha}=c^A_{q\alpha}+v^A_{q\alpha}+h^A_{q\alpha}$.  We
report known-label risk, coverage, and partial-identification bounds
\citep{imbens2004confidence}:
\begin{align}
C^{K,A}_{q\alpha}
&=\frac{c^A_{q\alpha}}{c^A_{q\alpha}+v^A_{q\alpha}},
\qquad
\gamma^A_{q\alpha}
=\frac{c^A_{q\alpha}+v^A_{q\alpha}}{N_{q\alpha}},\\
\left[C^{L,A}_{q\alpha},C^{U,A}_{q\alpha}\right]
&=\left[\frac{c^A_{q\alpha}}{N_{q\alpha}},
\frac{c^A_{q\alpha}+h^A_{q\alpha}}{N_{q\alpha}}\right].
\label{eq:matched-bounds}
\end{align}
Equation~\ref{eq:matched-bounds} reports what is established separately from what
unresolved labels could conceal.  When $c^A_{q\alpha}+v^A_{q\alpha}=0$, $C^{K,A}$ is undefined and $\gamma^A=0$.
The bounds are sharp under the recorded labels and have width $1-\gamma^A_{q\alpha}$.
Because $C^{U,A}$ is a fraction, inserting established-positive but irrelevant items
before the recall target can lower it without reducing the count of violations.  We
therefore interpret it only with feasibility and coverage and report non-dilutable
companions from the same prefix: the indicator of any known admissibility violation
and the known-violation count.  These companions measure observed violations rather
than unresolved worst cases.

To prevent infeasible routes from appearing artificially safe, we also define a
recall-constrained upper loss
\begin{equation}
\widetilde C^{U,A}_{q\alpha}
=
\begin{cases}
C^{U,A}_{q\alpha}, & F_{q\alpha}(\tau)=1,\\
1, & F_{q\alpha}(\tau)=0,
\end{cases}
\qquad
\mathcal{L}^{\mathrm{RC},A}_{\alpha}(\tau)
=
\left\langle
\widetilde C^{U,A}_{q\alpha}
\right\rangle_{\mathcal S},
\label{eq:route-loss}
\end{equation}
where $\mathcal S=\{\mathrm{RHELM},\mathrm{MemOps}\}$,
$\mathcal Q_s^+$ contains recall-evaluable queries from source $s$, and
$\langle X\rangle_{\mathcal S}=|\mathcal S|^{-1}\sum_s
|\mathcal Q_s^+|^{-1}\sum_{q\in\mathcal Q_s^+}X_q$ gives each source equal weight.
A historical non-usable composite is retained only for the frozen development
selector and provenance diagnostics in Table~\ref{tab:matched-prefix}; the primary
reported target is admissibility.  The loss is not
a conditional contamination probability: it combines a matched-prefix upper bound
with a unit infeasibility penalty, so recall, feasibility, bounds, and coverage remain
separate outputs.  The loss also decomposes into mean infeasibility
$\langle1-F_{q\alpha}(\tau)\rangle_{\mathcal S}$ plus the feasible-prefix contribution
$\langle F_{q\alpha}(\tau)C^{U,A}_{q\alpha}\rangle_{\mathcal S}$.  It is a comparative
route loss, not a safety certification or deployment acceptance threshold.
Conditional bounds average feasible queries within each source, whereas the second
loss term averages $F_{q\alpha}C^{U,A}_{q\alpha}$ over all recall-evaluable queries;
products of the displayed macro averages therefore need not reproduce that term.

\subsection{Exposure and disclosure}

For paired unit $i$ in relevance--admissibility cell
$c\in\{\mathrm{rel,adm},\mathrm{rel,inadm},\mathrm{irr,adm},
\mathrm{irr,inadm}\}$, let $D_i(1)$ and $D_i(0)$ denote literal target disclosure
when the focal candidate is exposed and withheld.  With $n_c$ paired units in cell
$c$, we define
the cell effect and reader selectivity gap as
\begin{align}
\widehat{\Delta}_c
&=\frac{1}{n_c}\sum_{i\in c}\left[D_i(1)-D_i(0)\right],\\
\widehat S
&=\widehat{\Delta}_{\mathrm{rel,adm}}-
\widehat{\Delta}_{\mathrm{rel,inadm}}.
\label{eq:paired-exposure}
\end{align}
A positive $\widehat S$ means that the exposed-minus-withheld disclosure contrast is
larger for relevant admissible evidence than for relevant inadmissible evidence;
$\widehat{\Delta}_{\mathrm{rel,inadm}}$ remains a separate outcome.  The next section
instantiates these record-, route-, and reader-level quantities on separate
benchmark-native and controlled populations.
\section{Evaluation Protocol}
\label{sec:evaluation-design}

We instantiate the framework along three questions: candidate support,
post-retrieval verification, and downstream enforcement (Table
\ref{tab:experiment-map}).  Distinct populations are never pooled; RHELM and MemOps
are combined only by a prespecified equal-source macro average, and the route-to-reader
evaluation is a frozen subset of that population.  Throughout, a \emph{verifier}
classifies candidate eligibility before prompting, a \emph{reader} generates the
answer, and a blinded \emph{judge} scores that answer.

\begin{table}[H]
  \centering
  \small
  \setlength{\tabcolsep}{4.5pt}
  \renewcommand{\arraystretch}{1.04}
  \caption{Experimental map.  Each row identifies the population, compared systems,
  and outcomes; rows are analyzed separately and never pooled.}
  \label{tab:experiment-map}
  \begin{tabular}{@{}>{\raggedright\arraybackslash}p{0.16\textwidth}
                       >{\raggedright\arraybackslash}p{0.40\textwidth}
                       >{\raggedright\arraybackslash}p{0.37\textwidth}@{}}
    \toprule
    Question & Population and comparison & Primary outcomes \\
    \midrule
    \textbf{RQ1: Support}
      & $3{,}767$ RHELM/MemOps queries; global, namespace, random,
      pre/post-filter, routed, and released-field arms; metadata stress
      & Recall, feasibility, bounds, coverage, constrained loss, typed exposure, work \\
    \textbf{RQ2: Verification}
      & 96 public-development queries with fixed top-20 pools plus
      $16$ controlled scenarios; fields versus text decisions
      & ROC-AUC, violation precision/recall, anchor false denial, consistency, overflip \\
    \textbf{RQ3: Enforcement}
      & $1{,}523$ paired benchmark-native reader cases plus
      $192$ controlled pairs per reader; no reader pooling
      & Answer correctness/quality, non-answer action, protected/stale disclosure,
      cell effects, selectivity \\
    \bottomrule
  \end{tabular}
\end{table}

\subsection{Datasets and splits}

The benchmark-native audit combines two complementary public benchmarks.
RHELM (Realistic, Heterogeneous, and Evolving Long-term Memory) provides
profile-conditioned temporal histories linked to heterogeneous sources,
whereas MemOps provides profiles and structured traces for remembering,
forgetting, updating, reflecting, and composed lifecycle operations
\citep{zhang2026rhelm,hao2026memops}. Their released evidence IDs,
provenance, operation targets, scopes, state transitions, and deterministic
operation replay provide required-evidence anchors and governance states
without subsampling or positive injection.

Development contains 23 disjoint persona/profile groups
($45{,}978$ memories; $1{,}104$ queries); evaluation contains 87 groups
($182{,}908$ memories; $3{,}767$ queries). RHELM candidates precede the
query date, whereas all method-visible MemOps memories enter global
support; namespace support restricts either pool to the query
persona/profile. RHELM retains 859 of 1,305 source pairs after requiring a
\texttt{conversation\_only} question and uniquely resolvable released
supporting references; the frozen group split assigns 523 of these pairs
to evaluation. Because usable anchors require positive scope, this study
tests fixed-budget reachability and robustness under trusted namespace
support, not whether namespace identity can be inferred from text.
Released-field eligibility rules are independent of routes and outcomes.

Two PhD-level annotators independently labeled 207 record--query pairs in
20 blinded packets. Krippendorff's $\alpha$, computed without adjudication,
was $.814$ for relevance, $.981$ for scope, $.670$ for lifecycle state,
and $.083$ for prohibited status. Relevance and scope were reproducible,
lifecycle agreement was moderate, and the sparse prohibited axis was not
independently reliable. Accordingly, prohibited status is interpreted as
released-field policy consistency rather than independently adjudicated
ground truth; omitting this axis preserves the direction of the principal
candidate-support results (Appendix~\ref{app:policy-sensitivity}).

The route-to-reader analysis uses all 523 evaluation RHELM cases and an
outcome-independent sample of $1{,}000$ from $3{,}244$ MemOps cases,
yielding $1{,}523$ cases across the same 87 groups, paired within each
reader. Readers receive only the question and ordered memory IDs and text
under frozen limits and are blinded to route identity and scores. Prompts
are deduplicated, with no output repair or selective rerun. An
outcome-independent 200-output sample audits judge agreement.

Two diagnostics isolate verification and exposure. A stable-hash
96-query public-development sample supplies 24 calibration and 72 analysis
cases. The constructed exposure study contains 16 scenarios, four in each
relevance--admissibility cell, with two paraphrased query pairs and three
fixed candidates per scenario. Neither diagnostic estimates population
prevalence.

\subsection{Interventions and comparison systems}

Dense arms use the pinned \texttt{Qwen/Qwen3-Embedding-8B} snapshot: 4,096-dimensional
float32 vectors, 512-token maximum, unit-normalized last-token pooling,
$s_{\mathrm{dense}}(m,q)=e_m^\top e_q$, and stable ID ties.  We compare global BM25,
global dense, BM25--dense fusion, recency+dense, namespace dense, current-only,
released-field, threshold, and cluster arms.  The released-field rule is a
label-aligned deterministic reference control, not an independently validated
governance predictor.  The post-hoc admissibility reanalysis evaluates the already
frozen global and namespace rankings across four
budgets; it changes scoring only, without reranking or evaluation-time tuning.
Policy-only, lifecycle-only, and combined released-governance filters are untuned.
Threshold and cluster routers use development-selected within-namespace settings and
a namespace-local dense fallback.  The threshold arm falls back for 96.5\% of
queries, while the cluster arm is practically indistinguishable from namespace dense;
their complete frozen definitions appear in Appendix~\ref{app:arms}.

Zero-call controls reuse the frozen embeddings and rankings: a ten-seed size-matched
random partition breaks provenance assignment, while global post-filtering at depths
20, 100, and 500 tests late deletion.  A separate policy-axis sensitivity omits only
the released policy predicate while preserving source anchors, scope, lifecycle,
routes, and the recall target.  None is retuned on evaluation.

Model roles are fixed.  GPT-5.6 Sol and Gemini 3.6 Flash are text-only verifiers (24
calibration; 72 analysis cases).  Controlled-exposure readers add DeepSeek V4 Pro,
with Claude Opus 5 as a separate unpooled replication.  Prespecified benchmark-native
readers are DeepSeek V4 Pro and Gemini 3.6 Flash; GPT-5.6 Luna is gated sequentially.
Claude Haiku 4.5 judges answers; GPT-5.1 audits 200 outputs.

\subsection{Metrics and analysis}

The preregistered development protocol fixed target recall at $\tau=.8$ before
evaluation and applied it unchanged to every arm.  It operationalizes a shared
utility floor: a route must recover at least 80\% of released required anchors before
its risk is compared.  Thus $.8$ is neither a retrieval-score threshold nor an
evaluation-tuned parameter.  For each recall-evaluable query we report evidence
recall and whether the route reaches $\tau$; on the smallest feasible prefix we report
known-label risk, label coverage, partial-identification bounds, and typed violations.
The loss $\mathcal{L}^{\mathrm{RC},A}$ in Equation~\ref{eq:route-loss} additionally
charges infeasible queries one.  The any-violation indicator and violation count
prevent dilution by inserted established-positive items.
For axis $j\in\{n,g,\ell\}$, typed exposure is
$V^{(j)}_{q\alpha}=|P_{q\alpha}|^{-1}\sum_{m\in P_{q\alpha}}
\mathbf 1\{x_{zj}(m)=0\}$.

Verifier diagnostics reported here include ROC-AUC, violation precision/recall,
and required-anchor false denial
$\operatorname{FD}=\sum_i d_i r_i/\sum_i r_i$.  Controlled studies add focal
consistency, stable overflip, and Equation~\ref{eq:paired-exposure}.  Downstream
outcomes are judged answer correctness, answer quality on a fixed 0--10 rubric
divided by 10, action-level non-answer,
and released-field protected/stale disclosure.  For outcome $j$, the natural reader analysis reports the
paired namespace-minus-global mean
$\Delta_j^{N-G}=\langle Y^{(j)}_{q,N}-Y^{(j)}_{q,G}\rangle_{\mathcal S}$ on its own evaluable
population.  Non-answer is not gold-action-aware over-refusal, and answer/disclosure
effects remain reader-specific.

Route settings were selected once on development from 165 candidates using
feasibility, historical non-usable loss, known-label risk, recall, work, and stable
setting ID.  The primary top-20 admissibility analysis only truncates those
frozen rankings: arms, ordering, and hyperparameters do not change.
Released-field attribution is post-hoc and untuned; the two-reader analysis was
prespecified, with GPT-5.6 Luna gated sequentially.  Corruption changes only
method-visible metadata at eight nested rates over ten seeds.  Weak-dominance breaks
are observed grid brackets, not interpolated thresholds.

Pointwise 95\% intervals use 10,000 percentile-bootstrap replicates, resampling
namespace groups for benchmark-native contrasts and prespecified strata otherwise;
reader comparisons are paired by namespace.  A zero-call sensitivity conditions
both global and namespace routes on the identical intersection of queries for which
each reaches $\tau=.8$.  Routes remain frozen, and models and
populations are not pooled.  A separate
observational GateMem audit uses a different population and is not combined with
these estimates \citep{ren2026gatemem}.

\section{Retrieval-Admissibility Verification Results}

\begin{figure}[!t]
  \centering
  \includegraphics[width=\textwidth]{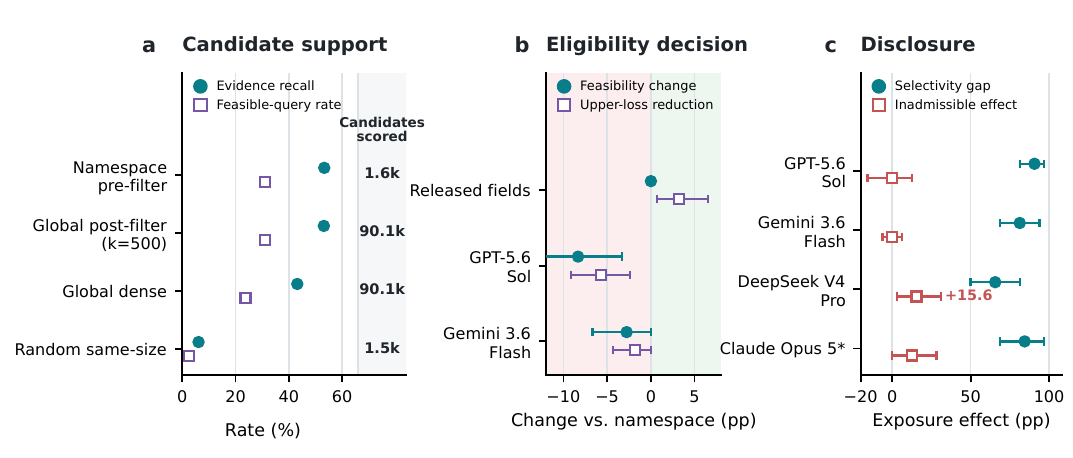}
  \caption{\textbf{Three separate evaluations yield different stagewise verdicts.}
  (a) Namespace beats random same-size support; late filtering retains global work.
  (b) The released-field reference preserves evidence, but the text-only gates do not.
  (c) Only DeepSeek V4 Pro's relevant-inadmissible contrast excludes zero; $^*$ marks
  the Claude Opus 5 replication.  Populations ($3{,}767$ queries, 72 verifier cases, and
  192 pairs per reader) are not pooled.}
  \label{fig:stagewise-evidence}
\end{figure}

Figure~\ref{fig:stagewise-evidence} summarizes three separate, unpooled evaluations of
candidate support, fixed-candidate verification, and controlled exposure.  Table~\ref{tab:main-comparison}
separately reports the same-population route-to-reader association.

\begin{table}[!t]
  \centering
  \scriptsize
  \setlength{\tabcolsep}{3.0pt}
  \renewcommand{\arraystretch}{1.00}
  \caption{\textbf{Frozen-route admissibility audit and downstream reader association.} (a) Post-hoc admissibility, top-20 results; rankings and settings are unchanged, coverage
  is .992; similarity evaluations count query--candidate score computations,
not wall-clock latency. (b) Equal-source reader averages and
  paired namespace-minus-global contrasts. (c) Typed matched-prefix exposure; axes
  can overlap.  Readers remain separate; $^{\dagger}$ marks a gated replication.}
  \label{tab:main-comparison}
  \textbf{(a) Candidate-support audit: absolute top-20 metrics}\par\vspace{2pt}
  \begin{tabular}{@{}>{\raggedright\arraybackslash}p{0.185\textwidth}
                         >{\raggedright\arraybackslash}p{0.205\textwidth}rrrrr@{}}
    \toprule
    Method & Visible information & Recall $\uparrow$ & Feasible $\uparrow$
      & $[C^{L,A},C^{U,A}]$ $\downarrow$
      & $\mathcal{L}^{\mathrm{RC},A}$ $\downarrow$ & \shortstack{Similarity\\evals. $\downarrow$} \\
    \midrule
    Global dense & text & .432 & .237 & [.187,.195] & .798 & 90,122 \\
    Global recency+dense & text + released order & .452 & .256 & [.198,.206] & .784 & 90,122 \\
    \textbf{Namespace dense} & text + trusted namespace & \textbf{.533} & \textbf{.311}
      & \textbf{[.118,.126]} & \textbf{.713} & \textbf{1,551} \\
    \midrule
    \multicolumn{2}{@{}l}{\textbf{$\Delta$ Namespace $-$ global}}
      & \textbf{+.101} & \textbf{+.074} & --- & \textbf{$-$.085}
      & \textbf{58.1$\times$ fewer} \\
    \bottomrule
  \end{tabular}

  \vspace{3pt}
  \textbf{(b) Downstream reader effects: namespace dense $-$ global dense}\par\vspace{2pt}
  \setlength{\tabcolsep}{2.4pt}
  \begin{tabular}{@{}lrrrr@{}}
    \toprule
    Reader & Accuracy $G\!\rightarrow\!N$ & $\Delta$ accuracy [95\% CI]
      & $\Delta$ quality [95\% CI] & $\Delta$ non-answer [95\% CI] \\
    \midrule
    DeepSeek V4 Pro
      & $.372\!\rightarrow\!.424$ & +.053 [+.020,+.084]
      & +.043 [+.023,+.062] & $-$.048 [$-$.068,$-$.031] \\
    Gemini 3.6 Flash
      & $.457\!\rightarrow\!.524$ & +.068 [+.047,+.090]
      & +.050 [+.036,+.065] & $-$.041 [$-$.057,$-$.025] \\
    GPT-5.6 Luna$^{\dagger}$
      & $.467\!\rightarrow\!.533$ & +.066 [+.039,+.096]
      & +.048 [+.031,+.066] & $-$.038 [$-$.059,$-$.015] \\
    \bottomrule
  \end{tabular}

  \vspace{3pt}
  \textbf{(c) Typed violation exposure in feasible matched prefixes}\par\vspace{2pt}
  \setlength{\tabcolsep}{5.0pt}
  \begin{tabular}{@{}lrrr@{}}
    \toprule
    Route & Wrong scope $\downarrow$ & Policy-disallowed $\downarrow$
      & Lifecycle-incompatible $\downarrow$ \\
    \midrule
    Global dense & .084 & .094 & .019 \\
    Namespace dense & .000 & .101 & .017 \\
    \bottomrule
  \end{tabular}
\end{table}

\paragraph{Candidate support.}

Namespace support raises top-20 recall .432 to .533 and feasibility .237 to .311
(Table~\ref{tab:main-comparison}a).  Both bound endpoints decrease from
[.187,.195] to [.118,.126] at .992
coverage.  At $K\in\{10,20,50,100\}$, paired recall and feasibility intervals exclude
zero; top-20 differences are +.101 [+.082,+.120] and +.074 [+.056,+.095].
Only .311 reach the .8 target.  Under the unit penalty in
Equation~\ref{eq:route-loss}, infeasibility contributes $-.074$ [95\% CI
$-.095,-.056$] and feasible-prefix risk $-.011$ [$-.016,-.007$] to the $-.085$
loss difference [$-.107,-.064$]; thus 87\% is attributable to feasibility.  Fixed-route sensitivity
over $\tau\in\{.5,.7,.8,.9,1\}$ preserves the namespace-versus-global direction.
On the 609 queries feasible under both routes, the upper bound likewise falls from
.1955 to .1191; the paired difference is $-.0763$ [95\% CI $-.0929,-.0600$]
(Table~\ref{tab:common-feasible}).

Hiding 20\% of evaluator judgments leaves routes and recall fixed but lowers
global/namespace coverage .992 to .786/.793 and widens bounds to
[.150,.364]/[.095,.302].

The effect is scope-specific: after excluding scope, upper risk is .126 for namespace
versus .123 for global.  Within feasible prefixes, any violation falls .528 to .396
and mean count 2.080 to 1.393; scope falls .084 to zero and lifecycle .019 to .017,
but policy rises .094 to .101 (Table~\ref{tab:main-comparison}c).  With prohibited-axis
$\alpha=.083$, policy is released-field consistency rather than adjudicated truth.  Omitting it preserves recall
(.432 to .533), feasibility (.237 to .311), and upper-loss reduction (.787 to .693)
($\Delta=-.0933$, 95\% CI [$-.115,-.071$]).

\begin{table}[H]
  \centering
  \scriptsize
  \setlength{\tabcolsep}{5.0pt}
  \renewcommand{\arraystretch}{1.00}
  \caption{\textbf{Support ablations and attribution controls.} Frozen top-20
  estimates at $\tau=.8$.  Random and anchor-preserving rows are ten-seed means;
  the latter uses released anchors. Similarity evaluations count query--candidate
score computations, not wall-clock latency.}
  \label{tab:support-ablation}
  \begin{tabular}{@{}lrrrr@{}}
    \toprule
    Support intervention & Recall $\uparrow$ & Feasible $\uparrow$
      & $\mathcal L^{\mathrm{RC},A}$ $\downarrow$ & \shortstack{Similarity\\evals. $\downarrow$} \\
    \midrule
    Global dense & .432 & .237 & .798 & 90,122 \\
    Random same-size & .061 & .024 & .978 & 1,540 \\
    Namespace pre-filter & .533 & .311 & .713 & 1,551 \\
    Global post-filter ($d=500$) & .531 & .311 & .713 & 90,122 \\
    Anchor-preserving same-size oracle & .840 & .747 & .390 & 1,551 \\
    \bottomrule
  \end{tabular}
\end{table}

Table~\ref{tab:support-ablation} separates support identity, filtering stage, and pool
size.  Random same-size support collapses recall, while post-filtering at depth 500
nearly matches namespace recall and loss but retains global scoring work.  The
anchor-preserving row is an unattainable diagnostic upper bound, not a deployable arm.

\paragraph{Eligibility verification.}

On 72 held-out development cases, the label-aligned reference lowers loss .032 with
unchanged recall/feasibility.  GPT-5.6 Sol and Gemini 3.6 Flash change
recall/feasibility/loss by $-.042/-.083/+.057$ and $-.014/-.028/+.018$, respectively.
Their ROC-AUC is .626/.522 and required-anchor false denial is .059/.020.  In the
separate controlled diagnostic, stable-admissible false denial is .266/.531/.594 for
GPT-5.6 Sol, Gemini 3.6 Flash, and DeepSeek V4 Pro.  At 1\% required-anchor false denial,
GPT-5.6 Sol detects nothing and no Gemini 3.6 Flash threshold qualifies
(Figure~\ref{fig:stagewise-evidence}b).

\emph{Metadata robustness.}  Policy/lifecycle/combined filters change recall/loss
$+.027/-.147$, $-.016/+.018$, and $+.010/-.124$.  Namespace denial, missingness, and
swaps first fail weak dominance at .20; false allow persists through .50 but exposes
.167 wrong-scope content.  These are grid brackets, not safety thresholds.

\paragraph{Route-to-reader association.}

On the same $1{,}523$ cases, every reader's judged-accuracy interval excludes zero
(Table~\ref{tab:main-comparison}b), as do five of six source-specific intervals.
Because namespace also raises recall, this observational association does not separate
admissibility from reachability; only the study below randomizes exposure.  A
200-output audit agrees strongly (correctness $\kappa=.728$; quality $\kappa_w=.884$;
protected/stale .975/.970).  Source-specific RHELM intervals, based on seven groups,
remain exploratory.

Filtering is not uniformly better: policy intervals cross zero; text verification
raises loss .0068 and lowers Gemini 3.6 Flash accuracy $-.0149$ [$-.0283,-.0011$];
metadata gates raise Gemini 3.6 Flash stale disclosure +.020 to +.024.  These results support the
namespace-restricted route comparison only; they do not show that filtering is
generally beneficial.

\paragraph{Controlled exposure.}

Across 16 scenarios, selectivity gaps for GPT-5.6 Sol, Gemini 3.6 Flash,
DeepSeek V4 Pro, and separately replicated Claude Opus 5 are .906/.813/.656/.844;
corresponding relevant-inadmissible literal-disclosure contrasts are
.000/.000/+.156/+.125.  Only DeepSeek V4 Pro excludes zero (95\% CI [.031,.312]).
These are reader-specific exact-marker effects over the 16 tested scenarios.

\section{Conclusion}

We presented a retrieval-admissibility verification framework with a query-conditioned
three-valued target and matched-recall bounds. It tracks memory IDs through prompt
exposure and links exposure to target-level answer disclosure. Trusted namespace support raised top-20
recall by .101 and feasibility by .074, eliminated wrong-scope matched-prefix exposure
under complete trusted labels, and was associated with reader-accuracy gains of
.053--.068 (plus .066 in the gated replication).  Evidence-preserving eligibility
checks and stagewise traces localize failures beyond recall or final-answer accuracy,
providing an auditable protocol for evaluating long-term-memory retrieval systems.

\clearpage

\bibliographystyle{plainnat}
\bibliography{references}

\appendix
\section{Declaration of LLM Use}
\label{app:llm-use}
We used OpenAI Codex and ChatGPT, as well as Anthropic Claude, as assistive
tools during the preparation of this work.  These tools supported code editing
and debugging, LaTeX preparation, and language polishing.  The research
questions, hypotheses, experimental protocols, dataset and model selection,
evaluation criteria, execution decisions, statistical analyses, interpretation
of results, and final claims were determined by the human authors.

All reported numerical results were produced by the frozen experimental
pipelines rather than generated by an LLM.  We manually reviewed the code,
figures, citations, experimental evidence, and final manuscript, and take full
responsibility for the submitted work.  LLMs used as experimental readers,
verifiers, or judges are part of the evaluated methodology and are documented
separately in the evaluation protocol and experimental appendices.

\section{Limitations and Broader Impact}
\label{app:limitations}

\paragraph{Interpretation and design implications.}
Retrieval admissibility asks whether the right evidence crossed each system boundary,
not only whether something useful was retrieved.  The audit separates unreachable
evidence, verifier over-denial, and disclosure after inadmissible exposure; coverage
and bounds keep sparse labels visible.  Trusted namespace support works by changing
reachability: correct provenance beats a same-size random partition, and late
filtering needs much deeper global retrieval.  Yet policy, lifecycle, and text-only
gates can also remove required evidence, so verification needs both eligibility-risk
and evidence-preservation axes.  A practical verifier should constrain candidate
support with authenticated provenance before semantic ranking, apply later gates only
when their fields are reliable, surface unresolved cases instead of imputing
permission, and retain auditable exclusion and exposure traces.

\paragraph{Limitations.}
The evidence covers two public sources and only seven RHELM groups; namespaces are
trusted rather than inferred.  One Claude Haiku 4.5 judge is shared, GPT-5.6 Luna is a sequential
replication, the alternate-judge audit covers 200 outputs, and fixed-candidate analysis
has 72 cases.  Human labels are a repeatability sample, with weak prevalence-corrected
agreement for prohibited status.  Source-execution hardware records are incomplete,
so candidate-count reductions are not wall-clock latency claims.  We do not establish
gold-aware over-refusal, internal use, a model ranking, safety certification, or
generalization beyond the tested corruption channels.  The frozen-route reader
comparison is associational; only the constructed exposure study randomizes prompt
exposure under its stated scenarios.  We also do not cover graph-structured selectors
in which unauthenticated structural writes can alter which authenticated records are
selected; our support result assumes flat candidate sets and trusted support labels
\citep{fei2026selectionintegrity}.

\paragraph{Broader impact.}
Record-level verification can reduce cross-principal, forgotten, or outdated evidence
exposure, but can also suppress legitimate history, amplify provenance errors, or
create refusal.  The framework therefore recommends auditable exclusion reasons,
unresolved status, evidence recall, exposure, utility, and recovery paths rather than
maximal filtering.  These safeguards matter when provenance or lifecycle metadata are
missing, corrupted, or controlled by an adversary.

\section{Metric Details}
\label{app:metric-details}

For status family $B\in\{U,A\}$, the matched-prefix quantities in
Equation~\ref{eq:matched-bounds} obey
\begin{align}
C^{K,B}_{q\alpha}&=\frac{c^B}{c^B+v^B}, &
\gamma^B_{q\alpha}&=\frac{c^B+v^B}{N},\label{eq:known-coverage}\\
C^{L,B}_{q\alpha}&=\frac{c^B}{N}=\gamma^B_{q\alpha}C^{K,B}_{q\alpha}, &
C^{U,B}_{q\alpha}&=\frac{c^B+h^B}{N}
=\gamma^B_{q\alpha}C^{K,B}_{q\alpha}+(1-\gamma^B_{q\alpha}).
\label{eq:partial-id}
\end{align}
Hence $C^{U,B}_{q\alpha}-C^{L,B}_{q\alpha}=1-\gamma^B_{q\alpha}$.  The bounds are obtained by
assigning every unresolved item first to the positive and then to the negative class;
if every item is unresolved, $C^{K,B}$ is undefined but the interval remains $[0,1]$.

The aggregate in Equation~\ref{eq:route-loss} decomposes exactly within each source.
Writing $\mathcal{L}^{\mathrm{RC},B}_{\alpha s}$ for the source-$s$ term, we have
\begin{equation}
\mathcal{L}^{\mathrm{RC},B}_{\alpha s}
=\underbrace{\frac{1}{|\mathcal{Q}^{+}_s|}\sum_{q\in\mathcal{Q}^{+}_s}(1-F_{q\alpha}(\tau))}_{\text{infeasibility}}
+\underbrace{\frac{1}{|\mathcal{Q}^{+}_s|}\sum_{q\in\mathcal{Q}^{+}_s}
F_{q\alpha}(\tau)C^{U,B}_{q\alpha}}_{\text{feasible-prefix contribution}}.
\label{eq:risk-decomposition}
\end{equation}
The reported $\mathcal{L}^{\mathrm{RC},B}_{\alpha}$ is the equal-source mean of
these terms.
Queries without known anchors are recall-unevaluable rather than assigned zero.

\paragraph{Retrieval and verifier diagnostics.}
Let $\mathcal Q_s^+$ be recall-evaluable queries from source $s$, with
$\mathcal S=\{\mathrm{RHELM},\mathrm{MemOps}\}$.  The equal-source macro operator,
evidence recall, feasibility, and typed violation rate are
\begin{align}
\langle X\rangle_{\mathcal S}
&=|\mathcal S|^{-1}\sum_{s\in\mathcal S}|\mathcal Q_s^+|^{-1}
  \sum_{q\in\mathcal Q_s^+}X_q,\\
\operatorname{Rec}_{q\alpha}(k)
&=\frac{|\{m_1,\ldots,m_k\}\cap\mathcal G_q|}{|\mathcal G_q|},\qquad
F_{q\alpha}=\mathbf 1\{\operatorname{Rec}_{q\alpha}(n_{q\alpha})\geq .8\},
\label{eq:retrieval-outcomes}\\
V^{(j)}_{q\alpha}
&=|P_{q\alpha}|^{-1}\sum_{m\in P_{q\alpha}}\mathbf 1\{x_{zj}(m)=0\},\qquad
x_{zj}\in\{n_z,g_z,\ell_z\}.
\label{eq:typed-violation}
\end{align}
Among $M$ verifier candidates, let indicators $d_i,k_i,v_i,r_i$ mark dropped,
known-label, known-violation, and required-anchor status.  ROC-AUC is evaluated on
known labels; filter precision, recall, and required-anchor false denial use their
displayed denominators:
\begin{align}
\operatorname{Prec}_{V}
&=\frac{\sum_i d_iv_i}{\sum_i d_ik_i},
\label{eq:verification-calibration}\\
\operatorname{Rec}_{V}
&=\frac{\sum_i d_iv_i}{\sum_i v_i}, &
\operatorname{FD}=\frac{\sum_i d_ir_i}{\sum_i r_i}.
\label{eq:filter-diagnostics}
\end{align}
For $J$ controlled focal pairs, let $\hat a^f_{\nu,+/-}$ be the focal argmax and
$\mathcal H_\nu$ its stable controls.  Focal consistency and stable overflip are
\begin{align}
\operatorname{FC}
&=J^{-1}\sum_{\nu=1}^{J}
  \mathbf 1\{\hat a^f_{\nu,+}=1\land\hat a^f_{\nu,-}=0\},\\
\operatorname{OF}
&=\frac{\sum_{\nu=1}^{J}\sum_{h\in\mathcal H_\nu}
  \mathbf 1\{\hat a^h_{\nu,+}\neq\hat a^h_{\nu,-}\}}
  {\sum_{\nu=1}^{J}|\mathcal H_\nu|}.
\label{eq:controlled-verification}
\end{align}
Unknown is a distinct argmax state; undefined denominators remain unavailable.

For the observational GateMem audit, let $X_{qa}$ indicate target exposure and
$Y_{qa}$ target disclosure.  For each checkpoint, form ordered discordant pairs
$D_q=\{(a,b):X_{qa}=1,\ X_{qb}=0\}$ and average first within checkpoint, then across
checkpoints:
\begin{equation}
d_q=\frac{1}{|D_q|}\sum_{(a,b)\in D_q}(Y_{qa}-Y_{qb}),\qquad
\widehat{\mathrm{RD}}=\frac{1}{|\mathcal{Q}_D|}
\sum_{q\in\mathcal{Q}_D}d_q.
\label{eq:discordant-rd}
\end{equation}
This prevents checkpoints with more route pairs from receiving more weight; it does
not make the route contrast causal.

For completeness, the released-field lifecycle rule used in
Section~\ref{sec:query-conditioned-admissibility} is
\begin{equation}
\ell_z(m)=
\begin{cases}
1, & s_m=\text{current},\\
1, & s_m\in\{\text{stale},\text{superseded}\},\ i_q=\text{history},\\
0, & s_m\in\{\text{stale},\text{superseded}\},\ i_q=\text{current},\\
?, & s_m=\text{unresolved}\ \text{or}\ i_q=\text{unresolved}.
\end{cases}
\label{eq:lifecycle}
\end{equation}
Writing $x_{zj}(m)$ for the recorded label on admissibility axis $j$, the verifier's
known exclusion reasons and trusted-namespace support are
\begin{align}
\operatorname{reason}_z(m)
&=\{j\in\{\mathrm{scope},\mathrm{policy},\mathrm{lifecycle}\}:x_{zj}(m)=0\},
\label{eq:reasons}\\
\mathcal{C}^{N}_{z\alpha}&=\{m\in\mathcal{M}:\widetilde n_{z\alpha}(m)=1\},\qquad
\pi^{N}_{z\alpha}=\operatorname{Rank}(q;\mathcal{C}^{N}_{z\alpha}).
\notag
\end{align}
The typed violation metric is defined in Equation~\ref{eq:typed-violation}.
Finally, for leakage, utility, or refusal outcome $j$ and its own eligible population
$\mathcal{Q}_j$, the strict-minus-base contrast is
\begin{equation}
\Delta_j^{S-B}=\frac{1}{|\mathcal{Q}_j|}\sum_{q\in\mathcal{Q}_j}
\left(Y^{(j)}_{q,\mathrm{strict}}-Y^{(j)}_{q,\mathrm{base}}\right).
\label{eq:safety-utility}
\end{equation}

For a matched prefix, a known-rate estimate is undefined when every status in its
chosen family is unresolved; the lower and upper bounds remain defined.  Queries with
no released evidence anchors are
excluded from recall averages, and the evaluable rate is reported separately.  Macro
averages never impute missing query metrics.  Duplicate ranked memory identifiers,
missing assessments, negative runtime measurements, and non-finite values are
rejected by the reference implementation.

\begin{table}[ht]
  \centering
  \small
  \caption{\textbf{Historical non-usable diagnostics within feasible matched-recall
  prefixes.} These source-macro $B=U$ quantities do not use the infeasibility penalty
  and are not pure wrong-scope measures.  Namespace support lowers the resolved rate
  and lower bound but also lowers coverage; the upper bound remains high.}
  \label{tab:matched-prefix}
\begin{tabular}{lrrrr}
\toprule
Support & Known non-usable & Coverage & Lower & Upper \\
\midrule
Global dense & 0.575 & 0.538 & 0.308 & 0.770 \\
Namespace dense & 0.348 & 0.355 & 0.134 & 0.778 \\
\bottomrule
\end{tabular}
\end{table}

Coverage differs sharply across status families because historical $B=U$ also
conjoins relevance, which is established only for released evidence IDs, whereas
$B=A$ uses governance fields available for almost every routed record.  Thus the
$B=U$ upper bound can remain high when namespace support lowers coverage; in the
main $B=A$ analysis, both bound endpoints decrease.

\section{Policy-Axis Sensitivity}
\label{app:policy-sensitivity}

The zero-call sensitivity in Table~\ref{tab:policy-sensitivity} reuses the exact
frozen top-20 routes and source-required anchors.  It omits only the released
policy-disallowed predicate; scope, lifecycle, recall target, and infeasibility
penalty are unchanged.

\begin{table}[H]
  \centering
  \scriptsize
  \setlength{\tabcolsep}{3.3pt}
  \caption{\textbf{The namespace result does not depend on the low-agreement policy
  axis.} Equal-source macro estimates at target recall .8.  For the policy-omitted
  loss difference, the paired 10,000-sample namespace-group bootstrap gives
  $-.0933$ [$-.115,-.071$].}
  \label{tab:policy-sensitivity}
  \begin{tabular}{@{}llrrrrr@{}}
    \toprule
    Axis family & Route & Recall & Feasible & $[C^L,C^U]$
      & $\mathcal L^{\mathrm{RC}}$ & Any violation \\
    \midrule
    Scope + policy + lifecycle & Global & .432 & .237 & [.187,.195] & .798 & .528 \\
    Scope + policy + lifecycle & Namespace & .533 & .311 & [.118,.126] & .713 & .396 \\
    Scope + lifecycle & Global & .432 & .237 & [.102,.111] & .787 & .356 \\
    Scope + lifecycle & Namespace & .533 & .311 & [.017,.025] & .693 & .134 \\
    \bottomrule
\end{tabular}
\end{table}

\begin{table}[H]
  \centering
  \small
  \setlength{\tabcolsep}{5.0pt}
  \caption{\textbf{The matched-prefix risk reduction persists on an identical
  feasible-query set.} Both routes reach target recall on the same 609 queries.
  Equal-source macro estimates use the full scope+policy+lifecycle target; the paired
  namespace-minus-global upper-bound difference is $-.0763$ [95\% CI
  $-.0929,-.0600$] from 10,000 namespace-group bootstrap replicates.}
  \label{tab:common-feasible}
  \begin{tabular}{lrrrr}
    \toprule
    Route & Known risk & Coverage & $[C^L,C^U]$ & Any violation \\
    \midrule
    Global dense & .1893 & .9918 & [.1873,.1955] & .5285 \\
    Namespace dense & .1122 & .9913 & [.1105,.1191] & .3721 \\
    \bottomrule
  \end{tabular}
\end{table}

\section{Blinded Dual-Rater Label Repeatability Audit}
\label{app:human-label-audit}

Two PhD-level annotators independently labeled the same 207 record--query pairs
from a source-stratified set of 20 blinded packets (10 RHELM and 10 MemOps).  The
audit was designed to cover both sources rather than estimate population-wide label
accuracy.  The audit packets omitted the
source-derived labels, reviewer identities were kept distinct, and both reviews were
locked before comparison.  No model-assisted labels entered the calculation and no
adjudication was applied.  Reviewers labeled relevance, scope, lifecycle state, and
prohibited status.  The legacy usable composite was derived deterministically from
the frozen conjunction of relevance, scope, and prohibited status; it predates the
lifecycle-aware target in Equation~\ref{eq:usable} and is not evidence that the full
target was independently elicited.

\begin{table}[H]
  \centering
  \small
  \setlength{\tabcolsep}{4.0pt}
  \renewcommand{\arraystretch}{1.08}
  \caption{\textbf{Scope and relevance labels are reproducible; lifecycle and
  prohibited status remain the harder axes.} Raw independent, unadjudicated agreement on the
  blinded human-audit sample.
  Uncertainty counts labels across both reviewers; nominal Krippendorff's $\alpha$
  excludes pairs containing an uncertain label.  The legacy usable composite is
  derived rather than independently elicited and excludes lifecycle.  Exact
  agreement with the deterministic released-field construction is reported as a
  consistency check, not external gold validation.}
  \label{tab:human-label-agreement}
  \begin{tabular}{@{}lrrrrr@{}}
    \toprule
    Axis & Exact agreement & Uncertain & Evaluable & $\alpha$ & Reviewer--source \\
    \midrule
    Relevance & 0.8937 (185/207) & 0 & 207 & 0.8135 & .797--.802 \\
    Scope & 0.9952 (206/207) & 0 & 207 & 0.9808 & .995--1.000 \\
    Lifecycle state & 0.8599 (178/207) & 23 & 184 & 0.6703 & .841--.947 \\
    Prohibited evidence & 0.8841 (183/207) & 0 & 207 & 0.0829 & .729--.826 \\
    Legacy usable composite (derived) & 0.9710 (201/207) & 0 & 207 & 0.9405 & .836--.855 \\
    \bottomrule
  \end{tabular}
\end{table}

The disagreement structure is concentrated rather than uniform.  Seventeen of 22
relevance disagreements exchange \emph{required} and \emph{supportive}; scope has one
disagreement; and 23 of 29 lifecycle disagreements exchange \emph{uncertain} and
\emph{current}.  All 24 prohibited-status disagreements have one reviewer marking
the item prohibited and the other marking it allowed.  Consequently, the audit
supports high inter-rater reproducibility for relevance and scope and shows that
experts can assess semantic usefulness separately from provenance eligibility on
this sample.  Lifecycle state is moderately repeatable.  Prohibited status is not
treated as independently reliable despite its high raw agreement, and the legacy
composite neither includes lifecycle compatibility nor validates the complete target
in Equation~\ref{eq:usable}.
Per-reviewer exact agreement with the deterministic released-field construction is
.797--.802 for relevance, .995--1.000 for scope, .841--.947 for lifecycle state,
.729--.826 for prohibited status, and .836--.855 for the legacy composite.  These
numbers check implementation consistency with released fields; because that
construction is not independent adjudicated truth, they do not convert the audit
into a validity study.

\section{Additional Expert-Audited Examples}
\label{app:verification-examples}

\begin{figure}[H]
  \centering
  \includegraphics[width=\textwidth]{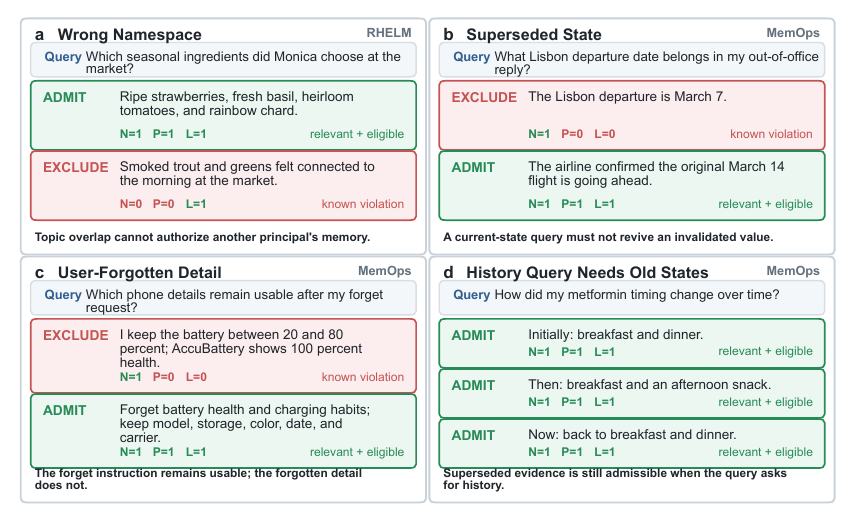}
  \caption{\textbf{Relevant records require query-conditioned eligibility checks.}
  Four abridged expert-audited examples underlying the admissibility taxonomy:
  wrong principal, superseded current state, explicit forgetting, and a historical
  query that requires superseded states.  Packet IDs, memory IDs, released labels,
  and verdicts are preserved in the example manifest; wording is shortened for
  legibility.  These cases illustrate coverage, not population prevalence.}
  \label{fig:verification-examples}
\end{figure}

\section{Closest-Work Comparison}

\begin{table}[H]
  \centering
  \scriptsize
  \setlength{\tabcolsep}{3.0pt}
  \renewcommand{\arraystretch}{1.08}
  \caption{\textbf{This work audits eligibility and downstream stages at the record
  level.} Relationship to nearby evaluation frameworks. ``Not primary'' describes
  each paper's reported estimands, not the capabilities of its system.  Our emphasis
  is one record-level audit across eligibility axes and stages.}
  \label{tab:closest-work}
  \begin{tabular}{@{}p{0.14\textwidth}p{0.22\textwidth}p{0.27\textwidth}
                       p{0.15\textwidth}p{0.14\textwidth}@{}}
    \toprule
    Work & Invalidity target & Primary unit / stages
      & Matched-recall bounds & Paired exposure estimand \\
    \midrule
    GateMem \citep{ren2026gatemem}
      & Principal access, updates, forgetting
      & Episode/checkpoint utility and leakage & Not primary & Not primary \\
    MemOps \citep{hao2026memops}
      & Operation target, scope, state transition
      & Structured operation trace and probes & Not primary & Not primary \\
    STALE \citep{chao2026stale}
      & Implicit invalidation of earlier state
      & State-resolution and answer probes & Not primary & Not primary \\
    A-TMA \citep{shi2026atma}
      & Current, historical, and transition state
      & Bank, retrieval, and answer diagnostics & Not primary & Not primary \\
    MemConflict \citep{tao2026memconflict}
      & Temporal, factual, and contextual conflicts
      & Retrieval, ranking, and final answers & Not primary & Not primary \\
    MemGate \citep{zhang2026memgate}
      & Query-conditioned contextual appropriateness
      & Learned admission gate, threat, and utility & Not primary & Not primary \\
    MRMS \citep{li2026mrms}
      & Eligibility, contradiction, supersession
      & Structured/vector/graph memory and gated projection & Not primary & Not primary \\
    MAP-Graph \citep{wang2026mapgraph}
      & Permission, provenance trust, revocation
      & Typed provenance graph, retrieval, and action gate & Not primary & Not primary \\
    RaMem \citep{yang2026ramem}
      & Context collapse and evidence validity
      & Anchoring, retrieval, and synthesis & Not primary & Not primary \\
    CI-Work \citep{fu2026ciwork}
      & Contextual information-flow violations
      & Task utility and privacy violation & Not primary & Not primary \\
    \textbf{This work}
      & \textbf{Scope, policy, and lifecycle compatibility}
      & \textbf{Same ID through exposure; target-level disclosure}
      & \textbf{Reported} & \textbf{Reported} \\
    \bottomrule
  \end{tabular}
\end{table}

\section{Constraint and Text-Only Inference Diagnostics}
\label{app:text-inference}

\begin{figure}[ht]
  \centering
  \includegraphics[width=\textwidth]{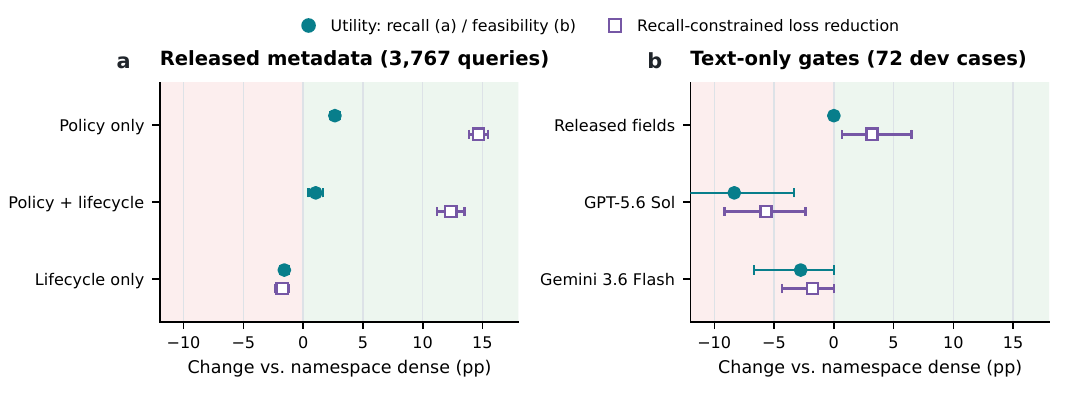}
  \caption{\textbf{More filtering is not monotonically better.} Changes from
  namespace dense in percentage points; right is better.  Circles show utility and
  open squares recall-constrained loss reduction for (a) released metadata on
  $3{,}767$ queries and (b) fixed-candidate gates on 72 development
  cases.  Whiskers are pointwise 95\% bootstrap intervals; verifiers remain
  separate.  Full corruption brackets appear in Table~\ref{tab:metadata-corruption}.}
  \label{fig:constraint-reliability}
\end{figure}

\begin{figure}[ht]
  \centering
  \includegraphics[width=\textwidth]{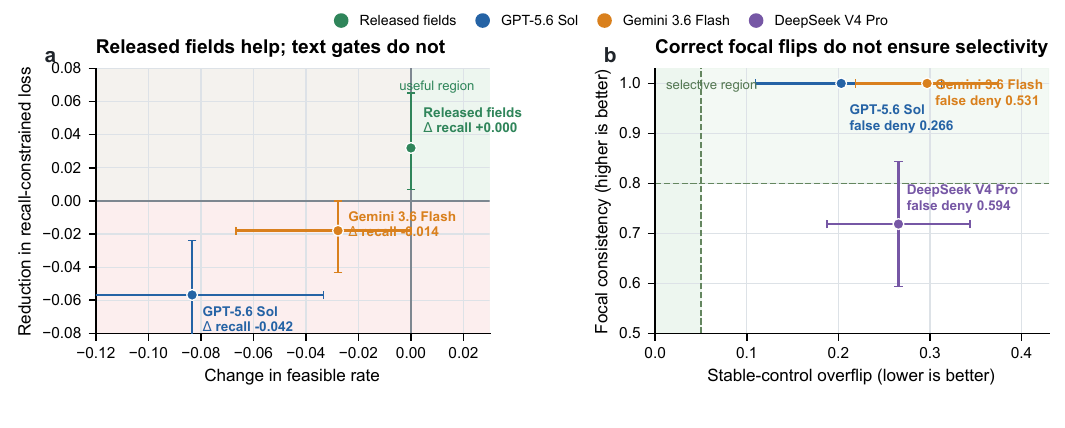}
  \caption{\textbf{Semantic verifiers recognize eligibility changes but do not apply them
  selectively.} (a) Feasibility and sign-reversed recall-constrained-loss differences
  versus namespace dense; labels give recall differences.  Whiskers are pointwise
  95\% intervals from 10,000 namespace-group bootstrap replicates stratified by
  source and query intent over 72 analysis cases.  (b) Controlled focal consistency
  versus stable-control overflip; labels report stable-admissible false-denial rate.
  The benchmark-native and controlled populations are separate and never pooled.}
  \label{fig:inference-gap}
\end{figure}

\begin{figure}[H]
  \centering
  \includegraphics[width=\textwidth]{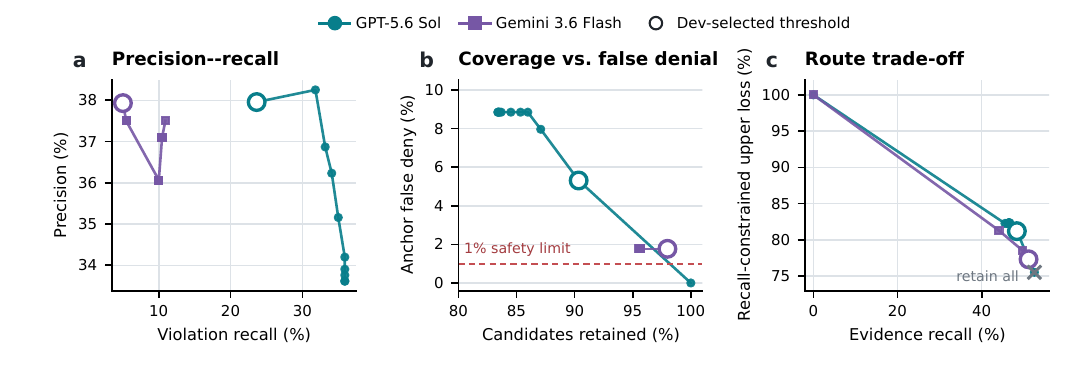}
  \caption{\textbf{No tested threshold satisfies the declared anchor-denial
  constraint while removing violations.} Thresholds span 0 to 1 in .05 steps;
  panels (a)--(b) show the practical
  $\geq .5$ region.  Open markers are the calibration-selected .95 thresholds.
  Candidate precision, violation recall, and anchor denial are analysis-candidate
  micro averages; route outcomes in (c) use the prespecified equal-stratum macro.
  GPT-5.6 Sol meets the 1\% false-denial constraint only by retaining every candidate
  and detecting no violations; Gemini 3.6 Flash has no tested point below the line.  The gray cross is the
  retain-all namespace route.}
  \label{fig:verifier-curves}
\end{figure}

\begin{table}[H]
  \centering
  \small
  \setlength{\tabcolsep}{4.0pt}
  \renewcommand{\arraystretch}{1.08}
  \caption{\textbf{Text-only verification does not yield a low-denial operating
  point.} Candidate-level diagnostics for the benchmark-native text-only verification
  experiment.  Values use the fixed development-selected operating points.  Higher
  ROC-AUC, violation precision, and violation recall are better; lower required-anchor
  false-denial rate is better.}
  \label{tab:text-inference-diagnostics}
  \begin{tabular}{lrrrr}
    \toprule
    Model & ROC-AUC & Violation precision & Violation recall & Anchor false deny \\
    \midrule
    GPT-5.6 Sol & $0.626$ & $0.350$ &
      $0.232$ & $0.059$ \\
    Gemini 3.6 Flash & $0.522$ & $0.679$ &
      $0.049$ & $0.020$ \\
    \bottomrule
  \end{tabular}
\end{table}

\section{Benchmark-Native Construction and Arms}
\label{app:arms}

\begin{figure}[H]
  \centering
  \includegraphics[width=\textwidth]{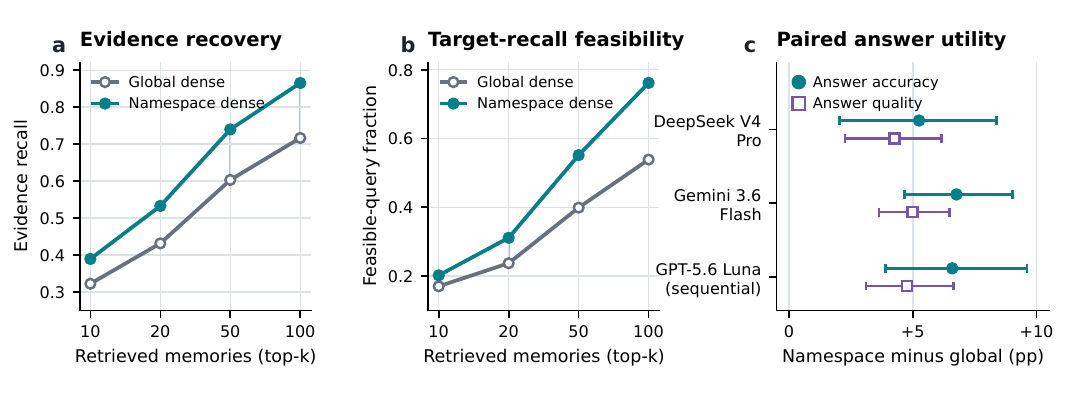}
  \caption{\textbf{Trusted namespace support improves retrieval and downstream
  utility.} Across $3{,}767$ queries, namespace raises (a) recall and (b) feasibility
  at every budget.  (c) All three readers gain accuracy on the frozen $1{,}523$ cases;
  whiskers are paired 95\% intervals.  GPT-5.6 Luna is sequential.  Claude Opus 5 was
  not executed as a reader on this population; Claude Haiku 4.5 served only as the
  shared blinded answer judge.}
  \label{fig:natural-closure}
\end{figure}

\begin{figure}[H]
  \centering
  \includegraphics[width=\textwidth]{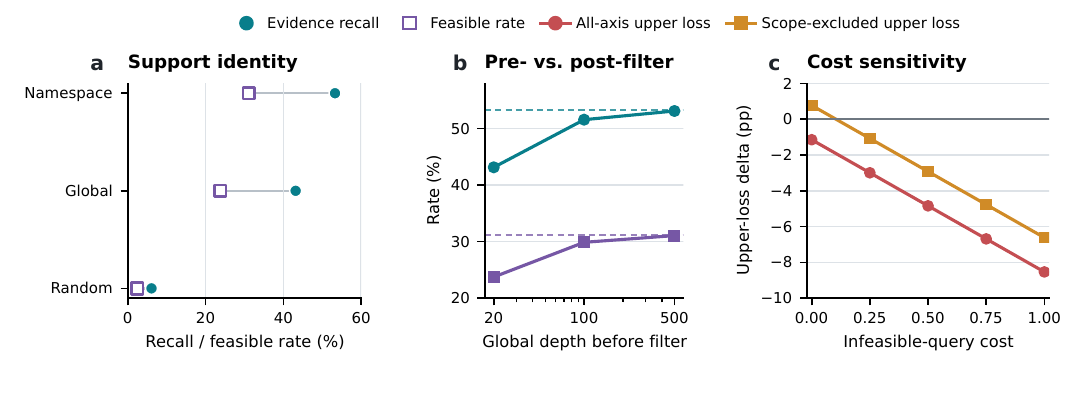}
  \caption{\textbf{Namespace support outperforms arbitrary candidate-count
  matching.} (a) The ten-seed random partition preserves
  source-level namespace-label counts but breaks provenance assignment; points show
  recall and feasible rate.  (b) Global top-$d$ is namespace-filtered and truncated
  to 20; dashed lines are namespace pre-filtering.  (c) Namespace-minus-global
  penalized risk as the explicit infeasibility cost varies; negative is better.
  Scope-excluded risk retains only released policy/lifecycle violations and unresolved
  labels.  Across all 87 groups, leave-one-group-out recall deltas are
  +.097 to +.108; the exact seven-group RHELM sign-flip $p$-value is .0156.}
  \label{fig:support-controls}
\end{figure}

\begin{figure}[H]
  \centering
  \includegraphics[width=\textwidth]{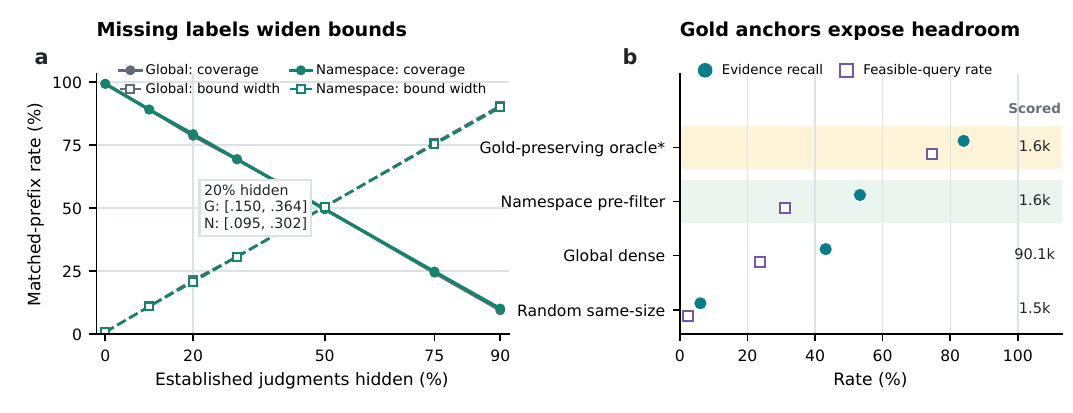}
  \caption{\textbf{Zero-call diagnostics expose uncertainty and retrieval
  headroom.} (a) Established composite evaluator judgments are hidden after
  retrieval, so routes, anchors, recall, and feasibility remain fixed.  Solid lines
  report known-label coverage and dashed lines the resulting
  partial-identification interval width; the callout gives lower and upper risk at
  20\% hiding.  (b) The gold-preserving arm retains every released required anchor
  and samples non-anchors to exactly the namespace support size.  Circles are recall,
  squares are feasible-query rate, and whiskers show one seed standard deviation
  where applicable.  \textsuperscript{*} The oracle reads released gold and is an attribution
  diagnostic, not a deployable router.}
  \label{fig:submission-diagnostics}
\end{figure}

\begin{table}[ht]
  \centering
  \scriptsize
  \setlength{\tabcolsep}{3.2pt}
  \renewcommand{\arraystretch}{1.08}
  \caption{\textbf{Namespace support improves answer utility for every frozen reader.}
  Benchmark-native same-population route-to-reader effects. Entries are namespace
  dense minus global dense with pointwise 95\% paired namespace-group bootstrap
  intervals.  Readers are reported separately and share one blinded Claude Haiku 4.5
  judge.  $^{\dagger}$GPT-5.6 Luna is a sequential robustness replication.}
  \label{tab:natural-end-to-end}
\begin{tabular}{lrrr}
\toprule
Reader & $\Delta$ answer accuracy & $\Delta$ answer quality & $\Delta$ non-answer rate \\
\midrule
DeepSeek V4 Pro & +0.053 [+0.020, +0.084] & +0.043 [+0.023, +0.062] & -0.048 [-0.068, -0.031] \\
Gemini 3.6 Flash & +0.068 [+0.047, +0.090] & +0.050 [+0.036, +0.065] & -0.041 [-0.057, -0.025] \\
GPT-5.6 Luna$^{\dagger}$ & +0.066 [+0.039, +0.096] & +0.048 [+0.031, +0.066] & -0.038 [-0.059, -0.015] \\
\bottomrule
\end{tabular}
\end{table}

\begin{table}[ht]
  \centering
  \scriptsize
  \setlength{\tabcolsep}{4.0pt}
  \renewcommand{\arraystretch}{1.05}
  \caption{\textbf{The direction is stable across sources, with wider RHELM
  uncertainty.} Source-specific namespace-dense minus global-dense effects in the
  benchmark-native reader analysis.  Intervals use 10,000 paired namespace-group bootstrap
  replicates.  The
  equal-source macro result in the main table is prespecified, but these rows expose
  source heterogeneity: DeepSeek V4 Pro accuracy on RHELM is the only accuracy interval that
  crosses zero.  Non-answer means $\mathbf{1}\{\mathrm{action}\neq\mathrm{answer}\}$,
  not gold-action-aware over-refusal.  RHELM intervals resample only seven evaluation
  namespace groups and should be read with that cluster count in view.}
  \label{tab:natural-source-effects}
  \begin{tabular}{@{}llrr@{}}
    \toprule
    Reader & Source & $\Delta$ answer accuracy & $\Delta$ non-answer rate \\
    \midrule
    DeepSeek V4 Pro & RHELM  & +.021 [$-$.037,+.075] & $-$.044 [$-$.075,$-$.016] \\
                    & MemOps & +.084 [+.052,+.116]   & $-$.053 [$-$.077,$-$.030] \\
    Gemini 3.6 Flash & RHELM  & +.059 [+.032,+.094] & $-$.042 [$-$.068,$-$.022] \\
                     & MemOps & +.076 [+.045,+.106] & $-$.039 [$-$.061,$-$.018] \\
    GPT-5.6 Luna$^{\dagger}$ & RHELM  & +.046 [+.004,+.094] & $-$.059 [$-$.096,$-$.016] \\
                              & MemOps & +.086 [+.053,+.121] & $-$.017 [$-$.034,$-$.001] \\
    \bottomrule
  \end{tabular}
\end{table}

\begin{table}[ht]
  \centering
  \scriptsize
  \setlength{\tabcolsep}{3.8pt}
  \renewcommand{\arraystretch}{1.05}
  \caption{\textbf{Case weighting preserves the namespace benefit.} Post-hoc
  case-weighted sensitivity for namespace dense minus global
  dense.  Each evaluable case receives equal weight across RHELM and MemOps; 95\%
  intervals resample namespace groups within source.  The prespecified equal-source
  macro remains primary, and readers remain separate.}
  \label{tab:natural-case-weighted}
  \begin{tabular}{@{}lrrr@{}}
    \toprule
    Reader & $\Delta$ answer accuracy & $\Delta$ answer quality
      & $\Delta$ non-answer rate \\
    \midrule
    DeepSeek V4 Pro
      & +.062 [+.033,+.092] & +.050 [+.033,+.068] & $-$.050 [$-$.069,$-$.032] \\
    Gemini 3.6 Flash
      & +.070 [+.048,+.094] & +.052 [+.037,+.067] & $-$.040 [$-$.057,$-$.024] \\
    GPT-5.6 Luna$^{\dagger}$
      & +.072 [+.045,+.101] & +.053 [+.036,+.071] & $-$.032 [$-$.049,$-$.014] \\
    \bottomrule
  \end{tabular}
\end{table}

RHELM development uses the first three persona-hash groups; evaluation uses the
remaining seven.  For each query, only conversation turns strictly earlier than its
date are candidates.  MemOps development uses the first 20\% of profile-hash groups;
evaluation uses the remaining 80 profiles.  Every method-visible dialogue memory in
the split is a global candidate.  MemOps has no cross-case wall clock, so its recency
arm uses released source order and is not a temporal-freshness claim.

All dense arms use \texttt{Qwen/Qwen3-Embedding-8B} at revision
\nolinkurl{1d8ad4ca9b3dd8059ad90a75d4983776a23d44af}, maximum length 512,
unit-normalized last-token pooling, float32 vectors, exact inner product, and stable
memory-ID tie breaking.  For unit vectors,
\[
s_{\mathrm{dense}}(m,q)=e_m^\top e_q.
\]
The hybrid arm uses reciprocal-rank fusion
\[
s_{\mathrm{RRF}}(m,q)=
\frac{1}{60+\operatorname{rank}_{\mathrm{BM25}}(m)}
+
\frac{1}{60+\operatorname{rank}_{\mathrm{dense}}(m)},
\]
and the recency control uses
$s_{\mathrm{rec}}(m,q)=s_{\mathrm{dense}}(m,q)+\gamma\rho(m)$, where
$\rho(m)\in[0,1]$ is normalized released order.

The threshold router incrementally assigns a memory to its most similar namespace
centroid when $\max_k\cos(e_m,c_k)\geq\theta$ and otherwise creates a cluster.  At
query time it selects at most $L$ centroids above $\theta$.  The cluster router uses
\[
\ell_k(x)=\gamma\log(n_k+\beta)+\frac{\cos(x,c_k)}{\tau},
\qquad
\ell_{\mathrm{new}}(x)=\log\alpha+
\frac{1-\max_j\cos(x,c_j)}{\tau_{\mathrm{new}}},
\]
assigns by the largest existing-or-new score, and routes over existing clusters.
An empty route in either router falls back to namespace-local dense retrieval.

\begin{table}[H]
  \centering
  \small
  \setlength{\tabcolsep}{4.0pt}
  \renewcommand{\arraystretch}{1.08}
  \caption{\textbf{All route settings were selected on development data only.}
  Frozen development-selected settings; all arms use $k=100$.}
  \label{tab:selected-settings}
  \begin{tabular}{lp{0.68\linewidth}}
    \toprule
    Arm & Selected setting \\
    \midrule
    Global BM25 & $k_1=1.5,\ b=0.75$ \\
    Global dense & Parameter-free \\
    Global BM25+dense RRF & $k_1=0.9,\ b=0.75,\ k_{\mathrm{RRF}}=60$ \\
    Global recency dense & $\gamma=0.1$ \\
    Namespace dense & Parameter-free \\
    Query-agnostic current-only & Remove stale and superseded records for every query \\
    Historical released-field v1 & Retain every namespace-local record for history;
      for current-state queries, remove released policy-disallowed, stale, and
      superseded records \\
    Threshold router & $\theta=0.80,\ L=3$ \\
    Cluster router & $\alpha=1,\ \beta=1,\ \gamma=0.5,\ \tau=0.1,\
      \tau_{\mathrm{new}}=0.4,\ L=1$ \\
    \bottomrule
  \end{tabular}
\end{table}

The artifact key \texttt{released\_governance\_oracle\_v2} denotes a corrected
post-hoc released-field rule, not a latent-label oracle, and is not one of the nine
development-selected arms.  It applies query-memory policy decisions to both history
and current-state queries, then applies lifecycle filtering only to current-state
intent.  All v2 attribution settings are fixed at top-100 and receive no tuning.

\begin{table}[H]
  \centering
  \scriptsize
  \setlength{\tabcolsep}{3.4pt}
  \renewcommand{\arraystretch}{1.08}
  \caption{\textbf{False-deny corruption is the earliest failure mode.} Observed
  metadata-corruption weak-dominance brackets over ten nested seeds.
  Namespace channels use global dense as the reference; policy, lifecycle, and intent
  channels use clean namespace dense.  Rates are tested grid points, not interpolated
  thresholds.  ``Not observed'' means no failure through the maximum tested rate .50.}
  \label{tab:metadata-corruption}
  \begin{tabular}{llll}
    \toprule
    Channel & Reference & Last dominant & First failure \\
    \midrule
    Namespace false deny & Global dense & .10 & .20 \\
    Namespace missing & Global dense & .10 & .20 \\
    Namespace source swap & Global dense & .10 & .20 \\
    Namespace false allow & Global dense & .50 & Not observed \\
    \addlinespace[2pt]
    Policy false deny & Clean namespace & .00 & .02 \\
    Policy false allow & Clean namespace & .30 & .40 \\
    Policy missing/fail open & Clean namespace & .30 & .40 \\
    \addlinespace[2pt]
    Lifecycle false stale & Clean namespace & .00 & .02 \\
    Lifecycle false current & Clean namespace & .50 & Not observed \\
    Lifecycle missing & Clean namespace & .50 & Not observed \\
    Query-intent flip & Clean namespace & .50 & Not observed \\
    \bottomrule
  \end{tabular}
\end{table}

\begin{table}[H]
  \centering
  \small
  \caption{\textbf{Namespace dense gives the strongest retrieval trade-off among the
  frozen routes.} All nine frozen v1 arms at $k=100$ on the source-macro public
  evaluation; main-text Table~\ref{tab:main-comparison}a truncates the same frozen
  rankings to top-20. The legacy
  penalized non-usable upper-risk field assigns 1.0 to infeasible or unresolved query outcomes.
  Candidates scored is a local diagnostic, not a production latency estimate.}
  \label{tab:all-arms}
\begin{tabular}{lrrrr}
\toprule
Arm & Recall & Feasible & \shortstack{Penalized non-usable\\upper risk} & Candidates \\
\midrule
Global BM25 & 0.448 & 0.284 & 0.913 & 90122 \\
Global dense & 0.717 & 0.539 & 0.873 & 90122 \\
Global BM25+dense RRF & 0.691 & 0.503 & 0.883 & 90122 \\
Global recency dense & 0.728 & 0.550 & 0.864 & 90122 \\
Namespace dense & 0.866 & 0.762 & 0.837 & 1551 \\
Query-agnostic current-only & 0.824 & 0.672 & 0.845 & 1549 \\
Historical released-field v1 & 0.869 & 0.763 & 0.828 & 1525 \\
Threshold router & 0.843 & 0.735 & 0.839 & 1470 \\
Cluster router & 0.865 & 0.761 & 0.837 & 1531 \\
\bottomrule
\end{tabular}
\end{table}

\begin{figure}[H]
  \centering
  \includegraphics[width=\textwidth]{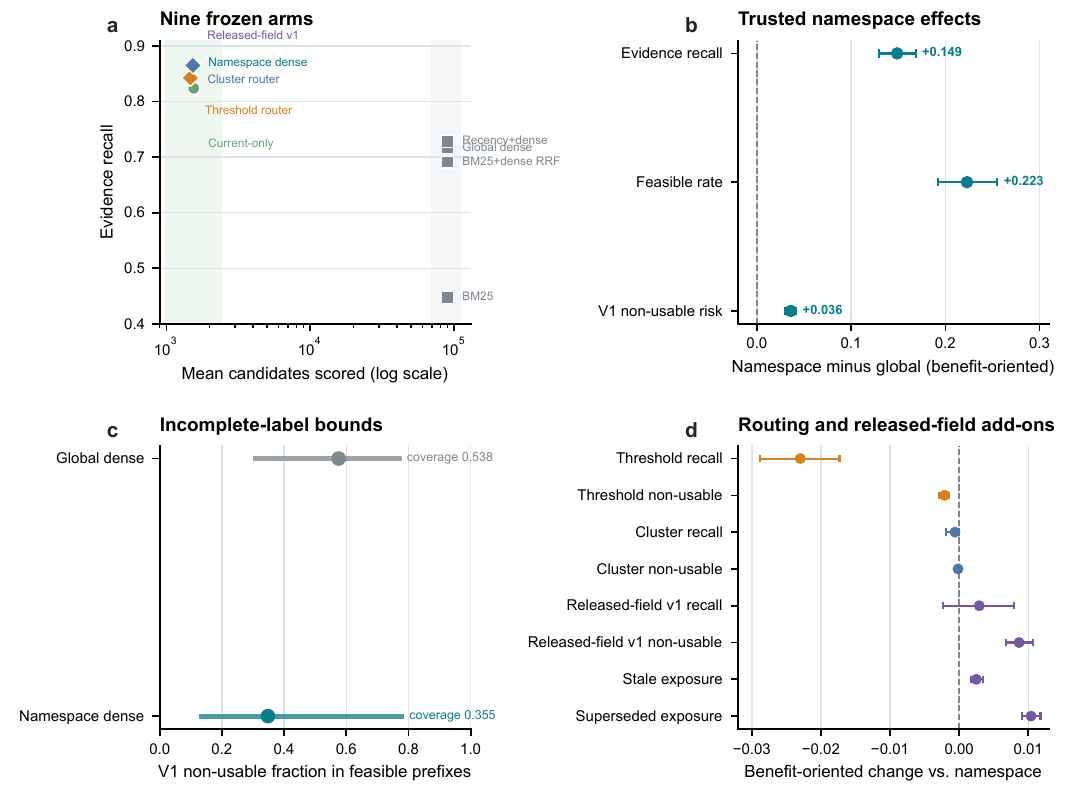}
  \caption{\textbf{Namespace support, rather than the tested routing refinements,
  separates the frozen retrieval regimes.} (a) Recall versus mean candidates
  scored for all nine arms.  (b) Namespace-dense minus global-dense effects under
  the historical non-usable loss.  (c) Conditional non-usable bounds and label
  coverage.  (d) Threshold, cluster, and historical released-field v1 differences
  versus namespace dense.  This figure preserves the original route semantics and
  cannot be read as the corrected v2 policy/lifecycle attribution.}
  \label{fig:retrieval}
\end{figure}

\section{GateMem Routing, Scoring, and Prediction}
\label{app:gatemem}

G0 is the exact upstream \texttt{rag\_naive} global TF--IDF route and G1 is the
upstream \texttt{rag\_policy} route with released policy logic.  G2 starts at the
authenticated asker, forms a connected component from explicit relationship IDs,
and admits chunks spoken by principals in that component; it reads neither query or
memory text nor leak targets when deciding support.  G3 adds only explicit pre-query
delete or supersede operations.  Because the release contains none, G3 is
trace-identical to G2.  G0 and G1 are upstream methods; G2 and G3 are diagnostics.

Each reader prompt contains the question and ordered retrieved record IDs and texts,
but not the arm name, scores, target strings, expected action, or gold answer.  Both
readers use temperature zero, JSON mode, and at most 384 output tokens.  The runs are
fresh, with no judge or output repair.  A strict parser accepts one object containing
action, answer, structured answer, and a string-list audit field; the audit field is
not scored, and parse failures receive conservative outcomes.

\begin{figure}[ht]
  \centering
  \includegraphics[width=\linewidth]{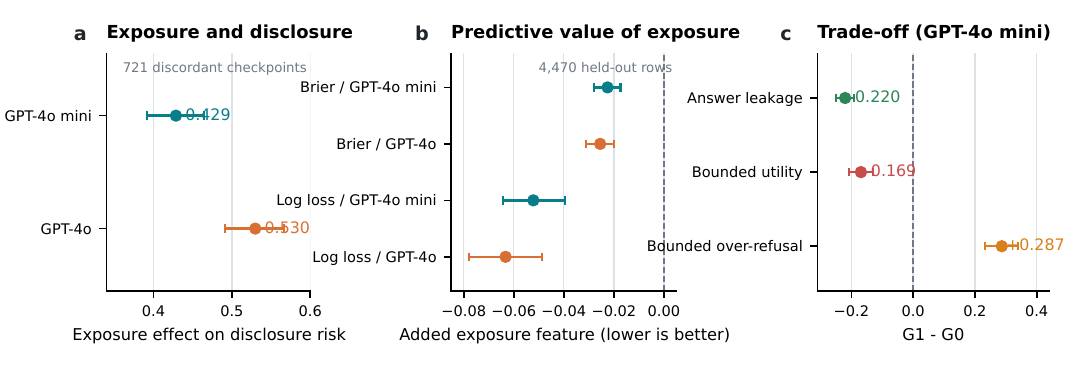}
  \caption{\textbf{Observed exposure predicts downstream disclosure but does not
  identify a causal effect.} (a) Exposed-minus-unexposed answer
  disclosure over $721$ discordant checkpoints.  (b) Predictive
  value of exposure on $4{,}470$ held-out rows.  (c) GPT-4o mini
  G1-minus-G0 leakage, bounded utility, and GateMem bounded over-refusal.  Readers are
  not pooled; route contrasts
  are non-causal and leakage and utility use different populations.  Exact reader
  snapshots are \texttt{gpt-4o-mini-2024-07-18} and \texttt{gpt-4o-2024-08-06}.}
  \label{fig:gatemem-observational}
\end{figure}

Privacy checkpoints use the released privacy context-leak and answer-leak rules;
safety checkpoints use the corresponding deletion rules.  The disclosure outcome
is the released rule-based end-to-end target indicator.  Utility correctness and
GateMem bounded over-refusal are rule-based auxiliary outcomes rather than LLM-judge
scores.
Parse failures are handled conservatively as leakage for privacy or safety and as
incorrect plus bounded over-refusal for utility.

The incremental predictor is logistic regression with $C=1$, the \texttt{lbfgs}
solver, and at most 2,000 iterations.  Categorical features are arm, domain, and
query type; additional baseline features are route non-emptiness and standardized
route width.  The augmented model adds only target exposure.  Each held-out episode
is scored by a model fit on all other episodes.  Brier and log-loss differences are
then resampled over episodes with 10,000 deterministic bootstrap replicates.

\section{Controlled Exposure Construction}
\label{app:paired-construction}

\begin{figure}[H]
  \centering
  \includegraphics[width=\textwidth]{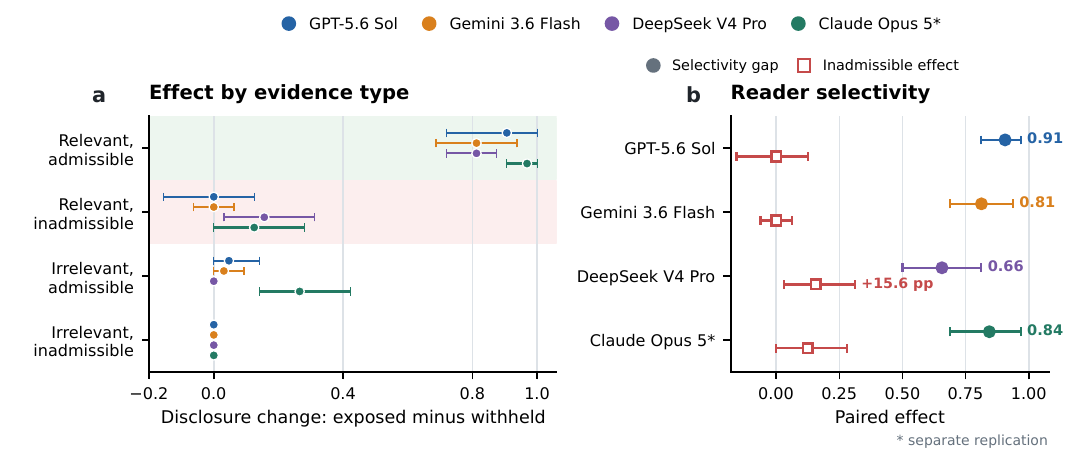}
  \caption{\textbf{Controlled exposure can become disclosure.} (a) Paired effects by
  evidence type; (b) selectivity gaps and relevant-inadmissible effects.  Whiskers are
  pointwise 95\% scenario-bootstrap intervals.  Readers are separate; Claude Opus 5
  is an unpooled replication, and DeepSeek V4 Pro's result is constructed, not a
  safety guarantee.}
  \label{fig:paired-exposure}
\end{figure}

Each of the 16 scenarios has two paraphrased query pairs, allow and block conditions,
and three manipulated roles: a relevant focal candidate, an irrelevant-admissible
candidate, and an irrelevant-inadmissible candidate.  The focal candidate is
admissible under allow and inadmissible under block.  For every
query--condition--candidate unit, two requests differ only in whether that candidate
is exposed.  This yields 192 paired units and 384 requests for each reader.

Every candidate carries a unique literal marker absent from the query and all other
candidates.  The normalized-substring scorer counts the marker even inside a refusal;
appearance after withholding is hallucinated disclosure.  Requests are stateless and
deterministically interleaved.  The original GPT-5.6 Sol, Gemini 3.6 Flash, and
DeepSeek V4 Pro readers and the
separately executed Claude Opus 5 replication each complete all 384 requests with one
attempt per request.  There is no transport retry, judge, output repair, selective
rerun, or model pooling.  The released result packages contain only content-free pair
scores, aggregates, hashes, costs, and bootstrap intervals.

\section{Mechanism Smoke}
\label{app:smoke}

The candidate-pool smoke uses four development packets and 16 evaluation packets,
with 7--12 candidate records per packet.  Namespace support preserves observed
recall at $1.0$ (global) and $1.0$ (namespace),
  changes measured wrong-scope leakage from $0.025$ to
  $0$, and changes the measured non-usable rate from
  $0.1189$ to $0.0824$.  This verifies
metric behavior under a known scope intervention; it is not a benchmark-corpus
performance estimate.

\begin{table}[ht]
\centering
  \small
  \setlength{\tabcolsep}{4.0pt}
  \renewcommand{\arraystretch}{1.08}
  \caption{\textbf{The scope intervention behaves as intended in a controlled smoke.}
  Candidate-pool results on 16 curated evaluation packets.}
\label{tab:smoke}
\begin{tabular}{lrrr}
\toprule
Support & Evidence recall & Wrong-scope leakage & Non-usable fraction \\
\midrule
Global dense & 1.0 & 0.025 & 0.1189 \\
Namespace dense & 1.0 & 0.000 & 0.0824 \\
\bottomrule
\end{tabular}
\end{table}

\section{Analysis Provenance}
\label{app:analysis-provenance}

Table~\ref{tab:analysis-provenance} separates prespecified analyses from corrections,
untuned attribution, sequential replication, and post-hoc robustness checks.  In
particular, v2 changes the status family used for scoring from historical non-usable
status ($B=U$) to admissibility ($B=A$); it does not alter rankings, selected
hyperparameters, or evaluation routes.  The released-governance attribution then
separates policy from lifecycle on those frozen rankings.

\begin{table}[H]
  \centering
  \scriptsize
  \setlength{\tabcolsep}{3.2pt}
  \caption{\textbf{The paper separates prespecified, corrected, replicated, and
  post-hoc analyses.} Provenance of the main analysis layers. ``Frozen route'' means that no
  ranking, threshold, or candidate order was changed for the analysis.}
  \label{tab:analysis-provenance}
  \begin{tabular}{@{}p{0.19\textwidth}p{0.19\textwidth}p{0.46\textwidth}@{}}
    \toprule
    Layer & Status & Relation to earlier evidence \\
    \midrule
    Retrieval setting selection & Prespecified development procedure
      & Selected before evaluation with historical $B=U$ loss; no test retuning. \\
    Frozen v1 evaluation & Held-out evaluation of the frozen v1 route family
      & Preserves the original non-usable status semantics and all nine arms. \\
    Admissibility v2 score & Post-hoc correction; frozen routes
      & Replaces $B=U$ by $B=A$ in scoring only; no reranking or retuning. \\
    Released-governance v2 & Post-hoc untuned attribution; frozen routes
      & Applies policy independently of intent and reports lifecycle separately. \\
    Two-reader analysis & Prespecified paired execution
      & Cases, routes, prompts, and two readers frozen before outcome scoring. \\
    GPT-5.6 Luna reader & Gated sequential replication
      & Repeats the three primary routes after the two-reader continuation criterion. \\
    Alternate judge & Post-hoc robustness audit
      & Outcome-independent 200-output sample; audits agreement without re-estimating effects. \\
    Support controls & Post-hoc zero-call diagnostic; frozen embeddings and labels
      & Permutes namespace assignment or changes only global pre-filter depth; no tuning or reranking model change. \\
    Policy-axis sensitivity & Post-hoc zero-call sensitivity; frozen routes
      & Omits only the released policy predicate; preserves anchors, scope, lifecycle, and recall target. \\
    Metric and verifier curves & Post-hoc zero-call sensitivity; frozen routes and scores
      & Varies recall target, infeasibility cost, or a fixed threshold grid without selecting a new operating point. \\
    Scenario robustness & Post-hoc zero-call sensitivity; frozen pair scores
      & Reports leave-one-scenario-out ranges and exact sign-flip tests without pooling readers. \\
    \bottomrule
  \end{tabular}
\end{table}

\subsection{Compute Resources and Execution Accounting}

Deterministic retrieval, scoring, bootstrap analysis, and artifact verification were
run on a single Windows 11 workstation with an AMD Ryzen 7 7800X3D CPU
(8 cores, 16 threads), 32\,GB system memory, and one NVIDIA GeForce RTX 5070 Ti
GPU with 16\,GB memory. The public verification and figure-generation path is
CPU-only and requires neither a GPU nor an API key. Historical embedding
materialization used Qwen3-Embedding-8B at the frozen revision reported in the
artifact; the 15.15\,GB model snapshot used CPU offload on the 16\,GB GPU.
All reader, verifier, and judge models were executed through provider-managed APIs.

Released receipts for the experiments reported in this manuscript account for 30,795
provider requests and USD~75.76. This comprises 13,326 GateMem reader calls (USD~23.35),
15,544 natural RHELM/MemOps reader and judge calls (USD~38.85),
192 text-verifier calls (USD~9.68), and 1,733 controlled exposure and verification
calls (USD~3.87). Receipts containing token accounting record at least 14.91 million
input tokens and 1.82 million output tokens; some historical natural-reader receipts
record calls and cost but not token totals. Where latency was recorded, median
per-call latency ranged from 0.98 to 20.00 seconds and the corresponding 95th
percentile ranged from 1.21 to 25.84 seconds. Provider wall time varies with rate
limits and concurrency. Total project-wide provider spend exceeded USD~500 after
including discarded pilots, partial or failed executions, and frozen exploratory
branches. Those charges are not presented as the cost of reproducing the reported
results because they are not fully attributable to the released evidence families.

\section{Reproducibility Artifact}

The \href{https://github.com/ziwang11112/right-memory-wrong-context}{\textcolor{blue}{\underline{public repository}}} releases the experimental code, frozen protocols, prompts,
synthetic fixtures, integrity tests, and machine-readable evidence derivatives used
in this study.  Hash-bound manifests connect each released evidence table to its
source snapshot and transformation script.  These materials support deterministic
offline regeneration of the reported aggregate tables, confidence intervals,
controls, and figures.  The released execution scripts also rebuild prompts from the
pinned public sources, call the configured providers with researcher-supplied API
credentials, normalize returned outputs, and regenerate the analyses under the
frozen protocols.  Credentials are never included in the artifact.

\end{document}